\documentclass[lettersize,journal]{IEEEtran}
\usepackage{amsmath,amsfonts}
\usepackage{algorithmic}
\usepackage{algorithm}
\usepackage{array}
\usepackage[caption=false,font=normalsize,labelfont=sf,textfont=sf]{subfig}
\usepackage{textcomp}
\usepackage{stfloats}
\usepackage{url}
\usepackage{verbatim}
\usepackage{graphicx}
\usepackage{cite}
\usepackage{booktabs}
\usepackage{xcolor}
\usepackage{ulem}
\usepackage{makecell} 
\begin{document}


\title{Funny-Valen-Tine: Planning Solution Distribution Enhances Machine Abstract Reasoning Ability}

\author{Ruizhuo Song, Member, IEEE,  Beiming Yuan
\thanks{This work was supported by the National Natural Science Foundation of China under Grants 62273036. Corresponding author: Ruizhuo Song, ruizhuosong@ustb.edu.cn}
\thanks{Ruizhuo Song and Beiming Yuan are with the Beijing Engineering Research Center of Industrial Spectrum Imaging, School of Automation and Electrical Engineering, University of Science and Technology Beijing, Beijing 100083, China (Ruizhuo Song email: ruizhuosong@ustb.edu.cn and Beiming Yuan email: d202310354@xs.ustb.edu.cn). }

\thanks{Ruizhuo Song and Beiming Yuan contributed equally to this work.}
}

\markboth{Journal of \LaTeX\ Class Files,~Vol.~14, No.~8, August~2021}%
{Shell \MakeLowercase{\textit{et al.}}: A Sample Article Using IEEEtran.cls for IEEE Journals}


\maketitle

\begin{abstract}
The importance of visual abstract reasoning problems in the field of image processing cannot be overstated.
Both Bongard-Logo problems and Raven's Progressive Matrices (RPM) belong to the domain of visual abstract reasoning tasks, with Bongard-Logo categorized as image clustering reasoning and RPM involving image progression pattern reasoning. This paper introduces a novel baseline model, Valen, which falls under the umbrella of probability-highlighting models. Valen demonstrates remarkable performance in solving both RPM and Bongard-Logo problems, offering a versatile solution for these reasoning tasks. 
Our investigation extends beyond the application of Valen, delving into the underlying mechanisms of probability-highlighting solvers. In revisiting how these solvers handle RPM and Bongard-Logo tasks, we realize that they approximate the solution to each reasoning problem as a distribution in which primary samples are compliant while auxiliary samples are not. This prompts us to propose that the learning objective of probability-highlighting solvers is not the distribution of correct solutions but rather one jointly delineated by primary and auxiliary samples. 
To bridge the discrepancies, we introduced the Tine method, an adversarial learning-based approach that helps Valen estimate a distribution close to that of the correct solutions. However, adversarial training in Tine suffers from instability. Motivated by this limitation, we model the sample distribution of reasoning problems as a mixture of Gaussian distributions, enabling Valen to capture the correct solution distribution more efficiently. This non-adversarial methodology leads to the development of the Funny method. Building on a similar Gaussian-mixture paradigm, we further propose the SBR method to plan the distribution of progressive pattern representations.
Overall, this paper contends that the key to enhancing solvers’ ability to address visual abstract reasoning problems lies in explicitly planning the distribution of predicted solutions to approach the correct solution distribution. {Codes are available in: https://github.com/Yuanbeiming/Funny-Valen-Tine-Planning-Solution-Distribution-Enhances-Machine-Abstract-Reasoning-Ability}
\end{abstract}


\section{Introduction}

\IEEEPARstart{D}{eep} neural networks have made remarkable achievements in various fields, including computer vision \cite{ImageNet, ResNet, A survey of convolutional neural networks, A survey of visual transformers}, natural language processing \cite{Transformer, GPT-3, Attention in natural language processing, survey of natural language processing}, generative models \cite{GAN, VAE, DiffusionModel}, visual question answering tasks \cite{CLEVERdataset}, and visual abstract reasoning \cite{RPM, Bongard1, Bongard2}.

Visual abstract reasoning plays a pivotal role in image processing. This task not only requires systems to detect objects and understand their relationships for logical inference, but also drives technological innovation and application expansion. By tackling such reasoning problems, we can stimulate the development of algorithms and technologies such as deep learning and graph convolutional networks, thereby advancing autonomous driving, medical diagnosis, security surveillance, and related fields. However, the field still faces insufficient and imbalanced data, limited model performance, and poor interpretability. Even large language models (LLMs), which have delivered breakthroughs in many domains, exhibit a significant performance drop on visual abstract reasoning tasks. Future research should therefore prioritize enhancing the abstract reasoning capabilities of machine intelligence and actively use such problems to evaluate the level of machine intelligence.

The RPM problem \cite{RPM} and the Bongard problem \cite{Bongard1,Bongard2} are two renowned examples within the realm of visual abstract reasoning. These two problems are the target problems that this paper aims to solve.

\subsection{RAVEN and PGM}

\begin{figure}[htp]\centering
	\includegraphics[trim=0cm 0cm 0cm 0cm, clip, width=8.5
 cm]{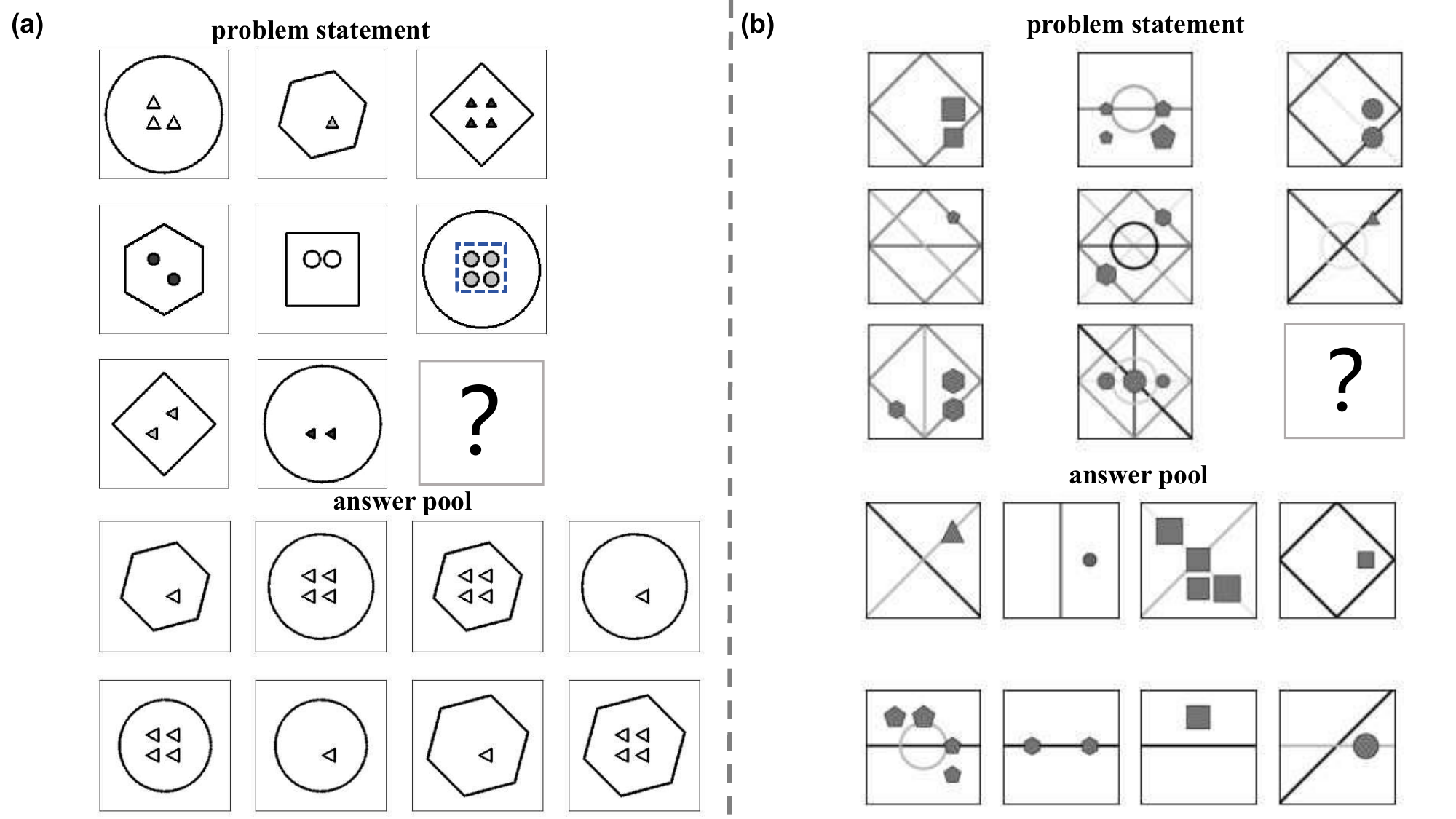}
	\caption{RAVEN instance (a) and PGM instance (b)}
\label{RAVEN}
\end{figure}

RPM\cite{RPM}, or Raven's Progressive Matrices, is a classic visual reasoning intelligence test typically presented as a 3$\times$3 matrix. Test-takers must observe and analyze visual features such as shape, size, direction, position, and their relations, identify the underlying rules, and select the missing element from the given option pool that completes the matrix. Thanks to its non-verbal format and proven utility in assessing general intelligence, RPM is widely used in psychology, education, and neuroscience as a tool for evaluating intellectual ability and investigating human cognitive development, brain function, and related processes. With advances in AI, researchers are now exploring algorithmic solutions to the RPM problem, though significant challenges remain
RAVEN\cite{RAVENdataset} and PGM\cite{PGMdataset} are classic RPM benchmarks; Figure \ref{RAVEN} illustrates sample instances. Intuitively, given the remarkable success of LLMs in many domains, one might expect them to handle such problems effortlessly. However, a series of studies~\cite{LLM1,LLM2,LLM3} has shown that LLMs fall far short on RPM tasks, underscoring the unique value of these benchmarks and indicating that abstract reasoning remains a scarce and essential capability for machine intelligence.

\subsection{Bongrad-Logo}

\begin{figure}[htp]\centering
	\includegraphics[trim=0cm 0cm 13cm 0cm, clip, width=5
 cm]{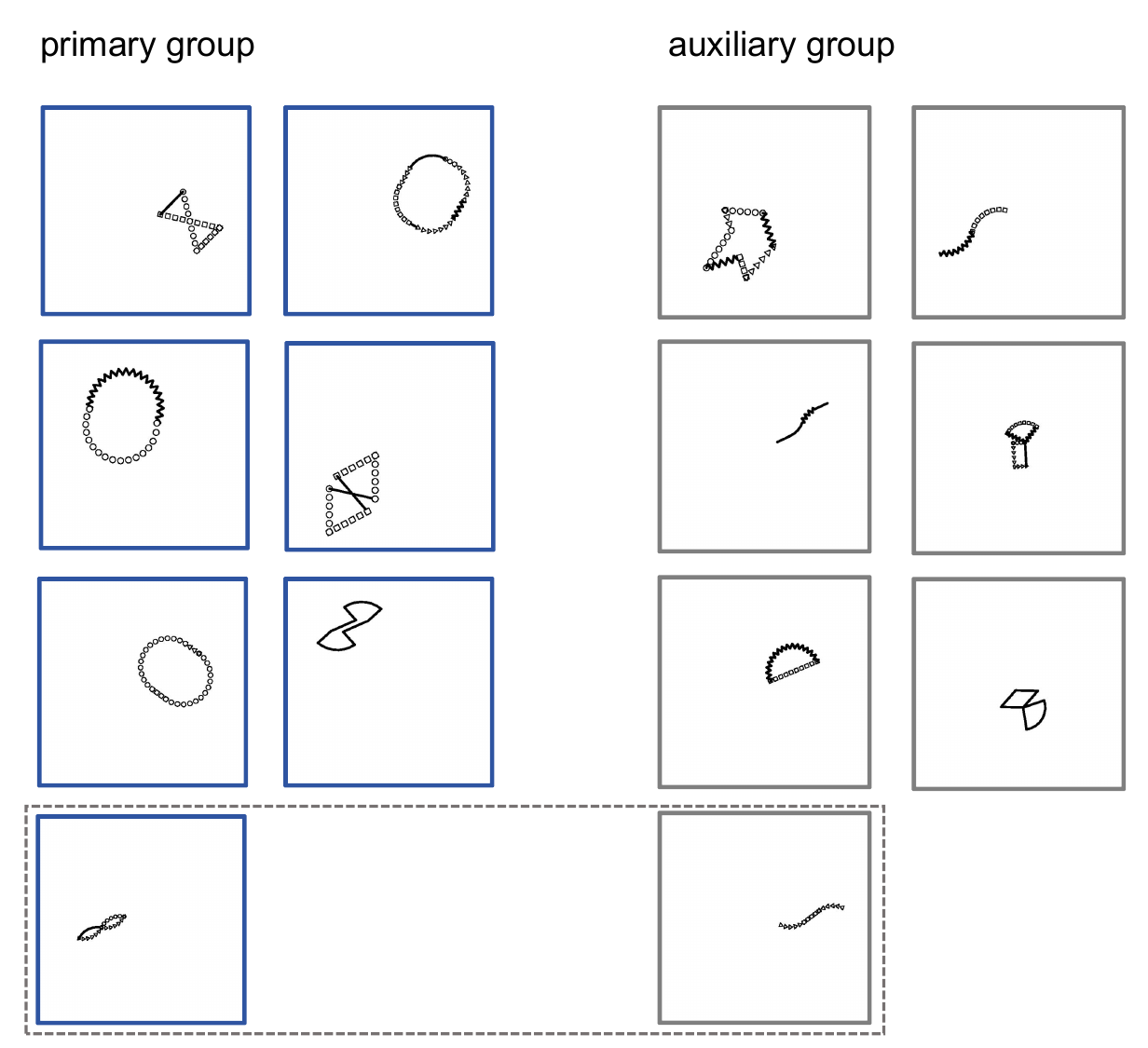}
	\caption{Bongard-Logo instance}
\label{Bongard}
\end{figure}

Distinct from RPM tasks, the Bongard problem \cite{Bongard1, Bongard2} belongs to image clustering \cite{clustering} reasoning problems. It typically involves multiple images divided into primary and auxiliary groups: images in the primary group share an abstract concept governed by specific rules, whereas images in the auxiliary group violate these rules. Solving Bongard problems demands that deep learning algorithms accurately classify the remaining ungrouped images. As a specific instance within abstract reasoning, the Bongard-Logo problem \cite{Bongard2} poses a significant challenge due to its inferential complexity. Each case comprises 14 images: six from the primary group, six from the auxiliary group, and two ungrouped images to be classified. The grouping criterion relies on geometric shapes and their arrangement. Figure \ref{Bongard} illustrates one such case.

Bongard-logo problems are categorized into four sets: The Free-form Shape Test Set (FF) aims to evaluate a model's ability to generalize to more complex free-form shape concepts. The Basic Shape Test Set (BA) assesses the model's generalization capability to new combinations of basic shape concepts. The Combinatorial Abstract Shape Test Set (CM) tests the model's abstract understanding of novel attribute combinations, where all individual attributes have been seen in the training set, but the combinations are entirely new. The Novel Abstract Shape Test Set (NV) evaluates the model's ability to discover new abstract concepts, particularly by withholding one attribute and all its combinations to examine the model's extrapolation capability. Solvers are trained on the training set and tested on these four sets to evaluate their intelligence levels from multiple dimensions.

\section{Related work}

\subsection{RPM solver}

{Discriminative models for image reasoning, such as CoPINet \cite{CoPINet}, LEN+teacher \cite{LEN}, and DCNet \cite{DCNet}, offer diverse solutions, focusing on learning differences and latent rules. Connectionist models: SCL \cite{SCL}, SAVIR-T \cite{SAVIR-T}, and neural-symbolic systems: PrAE \cite{PrAE}, NVSA \cite{NVSA}, ALANS \cite{ALANS} incorporate various techniques to improve reasoning accuracy and interpretability. By combining effective methods, RS-TRAN \cite{RS} have achieved impressive results on the RPM problem. Triple-CFN implicitly extracts information about concepts and reasoning units, and ``indexes'' this reasoning-unit information according to the concepts, achieving noteworthy reasoning accuracy. CRAB \cite{CRAB}, based on a Bayesian modelling approach, has established a highly customised, controlled sandbox that aligns seamlessly with its proprietary methodology, embodied in the creation of a new RAVEN database. This bespoke environment, although it forfeits inherent core challenges of RAVEN such as solution diversity and uncertainty, has enabled CRAB to achieve remarkable breakthroughs. The community looks forward to assessing the broader implications of this approach for future research. DRNet \cite{aaai} employs parallel CNN and ViT to extract local and spatial features, and uses a learnable fusion module to model abstract rules explicitly, offering a concise new framework for visual abstract reasoning. NCD \cite{NCD} converts the unsupervised RPM task into contrastive learning via pseudo labels, and employs negative answer sampling plus decentralized features to adaptively cancel noise and inter-problem centroid shifts, enabling automatic rule representation extraction without any annotation and offering a scalable new paradigm for unsupervised abstract reasoning.}

\subsection{Bongard solver}

{Researchers have tackled Bongard problems along three main avenues: symbolic models, convolutional neural networks, and synthetic-data strategies. Depweg et al. combined a formal language with symbolic vocabulary and Bayesian inference, yet their method struggled with complex instances and scaled poorly \cite{Bongard1}. Kharagorgiev and Yun pre-trained CNNs on image datasets for feature extraction \cite{Bongard3}. Nie et al. experimented with CNN-based meta-learning, but results remained sub-optimal \cite{Bongard2}. These approaches exhibit distinct strengths and weaknesses, underscoring the need for further research. Yuan proposed PMoC \cite{PMoC}, a probabilistic model that measures how likely auxiliary group samples fit the main group’s distribution—an approach that has shown strong performance. Notably, Triple-CFN \cite{Triple-CFN} has successfully applied a unified framework to both Bongard-Logo and RPM problems.}

\section{Methodology}

Both Bongard-Logo problems and Raven's Progressive Matrices (RPM) fall under the umbrella of visual abstract reasoning tasks. Specifically, Bongard-Logo problems can be categorized as image clustering reasoning, while RPM challenges involve image progression pattern reasoning.
We designed a general solver, Valen, for both RPM and Bongard-Logo problems. Furthermore, to enhance a solver's ability to tackle visual abstract reasoning tasks, we propose that its solution distribution should be explicitly planned to conform more closely to that of correct solutions. Therefore, we developed three methods to perform this explicit planning: Funny, Tine and SBR.

Here we clarify the definitions: ``primary samples" are those conforming to the core patterns of reasoning problems (e.g., correct RPM options or primary group samples in Bongard-Logo instances); ``auxiliary samples" are those that do not (e.g., incorrect RPM options or auxiliary group samples in Bongard-Logo problems).

\section{Valen: a new baseline model for RPM and Bongard-Logo problems}

In this section, we introduce Valen (Visual Abstraction Learning Network), our baseline model designed to address both RPM, an image progression pattern reasoning problem, and Bongard-Logo, an image clustering reasoning problem.

\subsection{Valen targeting RPM problem}

The RPM problem challenges participants to complete a 3$\times$3 image matrix by selecting an answer from an option pool. The correct option is the one that preserves the progression patterns reasoned from the incomplete matrix (i.e., the problem statement) \cite{RPMInductivebias}. 
To this end, we introduce a baseline named Valen: it assesses the plausibility of the completed 3$\times$3 matrix based on its internal progressive patterns and outputs a corresponding probability.

\begin{figure}[htp]\centering
	\includegraphics[trim=0cm 0cm 0cm 0cm, clip, width=8.5
 cm]{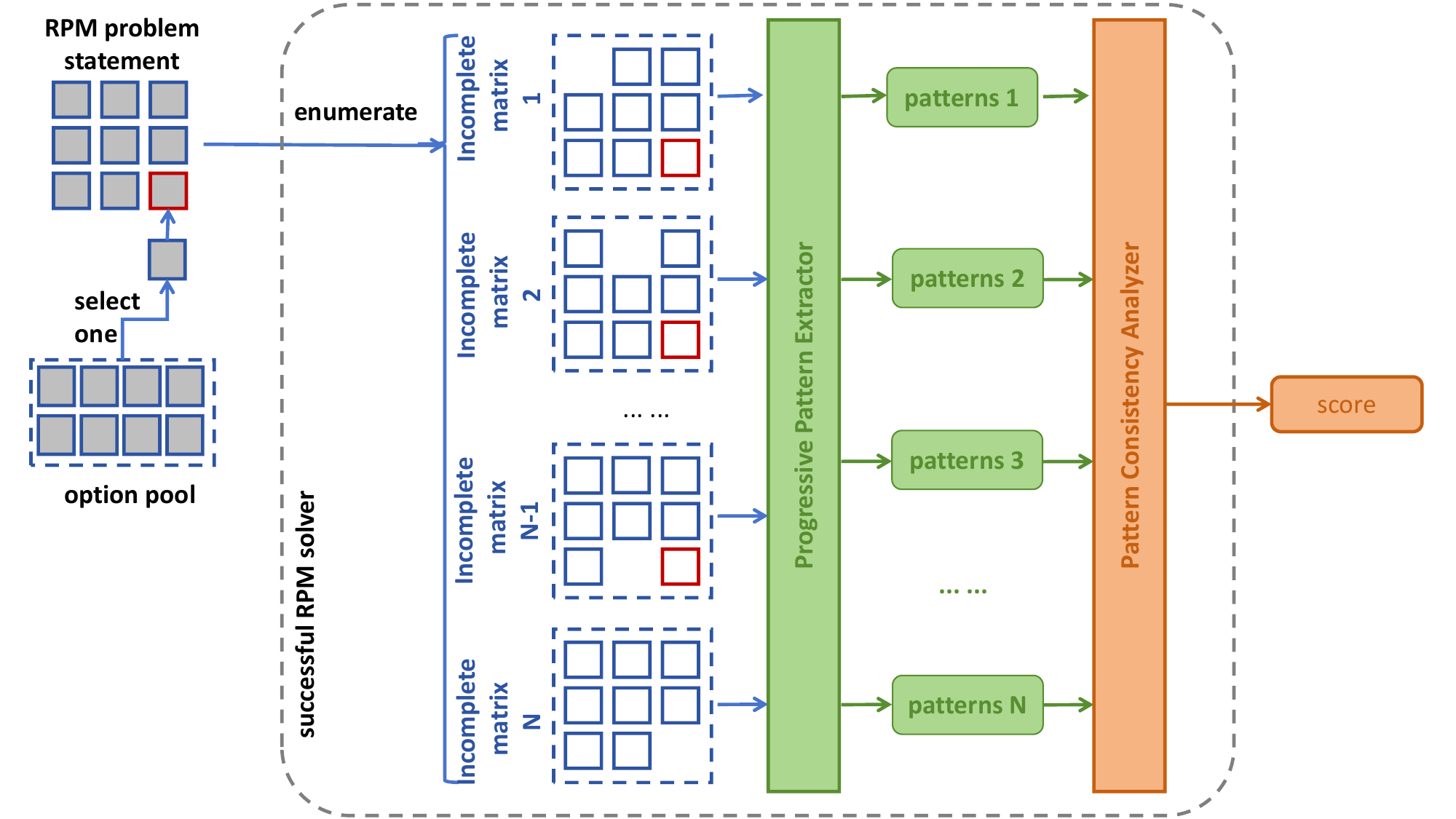}
	\caption{The reasoning process of an effective RPM problem solver}
\label{The reasoning process of an envisioned RPM problem solver}
\end{figure}

A defining feature of RPM problems is that an incomplete matrix (missing one image)  still retains sufficient information to reason about the complete progressive patterns. Based on this property, we argue that the reasoning process of an effective RPM problem solver should align with the illustration in Figure \ref{The reasoning process of an envisioned RPM problem solver}.
As seen in the figure, our envisioned RPM solver consists of two modules: the Progressive Pattern Extractor and the Pattern Consistency Analyzer. Additionally, our envisioned RPM solver can perform the following three reasoning processes: 
1. The process indicated by the blue line: When an option from the pool completes the RPM problem statement into a full $3\times 3$ progression matrix, the solver can enumerate all $N$ possible incomplete versions of the completed 3$\times$3 matrix. 
2. The process denoted by the green line: For each enumerated incomplete matrix, the Progressive Pattern Extractor extracts its progressive pattern, yielding $N$ sets of progressive patterns. 
3. The process marked by the orange line: The solver utilizes its Pattern Consistency Analyzer to evaluate the consistency of these $N$ sets of progressive patterns and outputs a score to quantify this consistency. This score also serves as the score for the corresponding option.

Since this paper provides a clear definition of the reasoning process for a successful RPM solver, the Valen model is derived by refining these three steps. Specifically, Valen comprises three stages: (i) representation extraction of matrices and enumeration of incomplete matrices, (ii) extraction of progressive patterns from incomplete matrices, and (iii) consistency analysis of progressive patterns. We describe each stage individually.

\subsubsection{Representation extraction of matrices and enumeration of incomplete matrices}
\begin{figure}[htp]\centering
	\includegraphics[trim=0cm 0cm 0cm 0cm, clip, width=8.5
 cm]{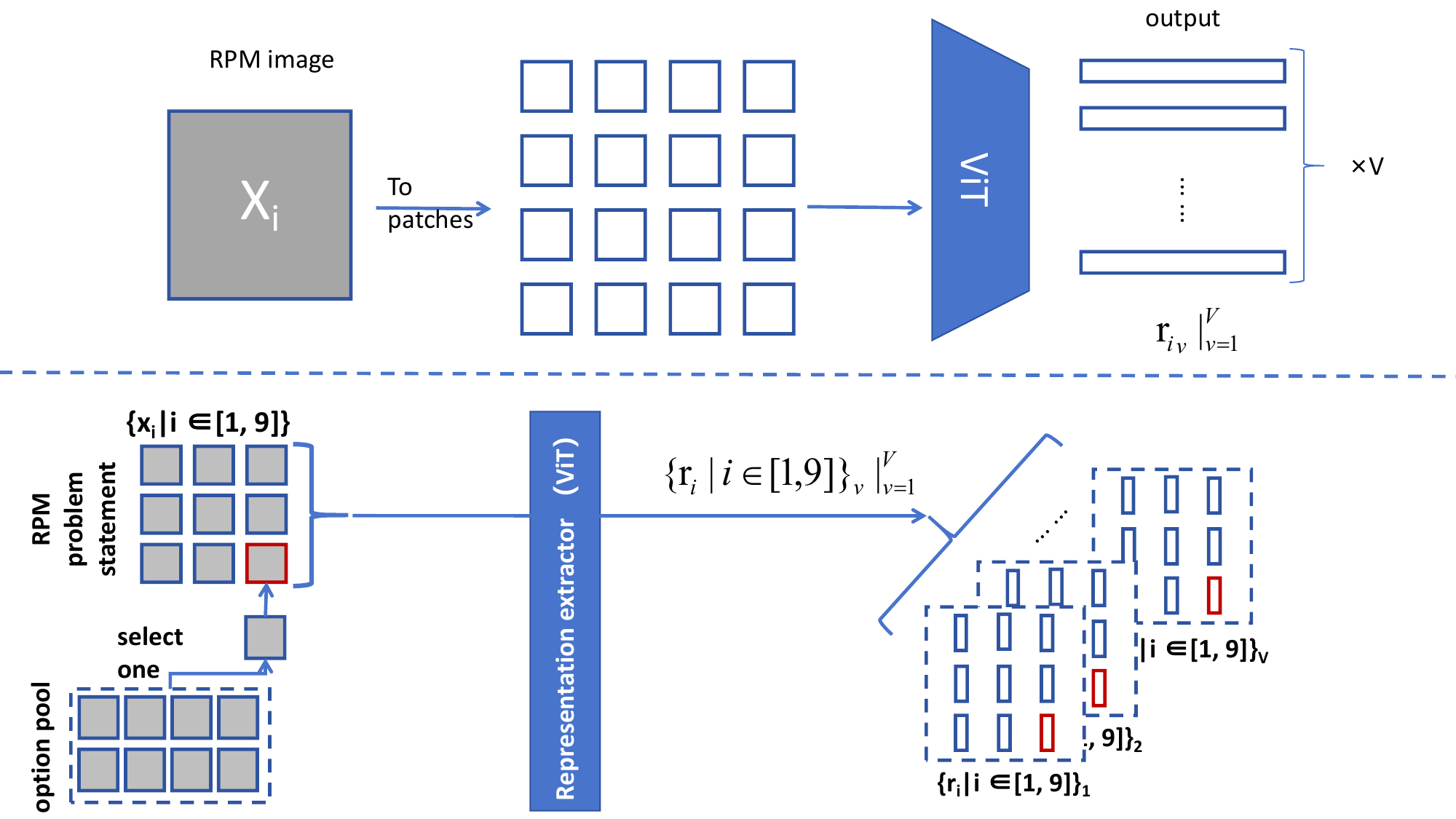}
	\caption{Representation extraction of matrix}
\label{representation extraction of matrix}
\end{figure}

This process in Valen follows the structure illustrated in Figure \ref{representation extraction of matrix}.
As shown in this figure, when the problem statement of the RPM task is completed by a specific option to form a $3\times 3$ progression matrix, denoted as $\{x_i | i \in [1,9]\}$, Valen utilizes a basic Vision Transformer (ViT) as a feature extractor to obtain multi-viewpoint representations of $\{x_i | i \in [1,9]\}$. These representations are denoted as $\{r_i | i \in [1,9]\}_v$, where $v \in [1, V]$, $V$ is the total number of viewpoints, $v$ represents the index of the viewpoint, and $i$ represents the index of the image in the matrix.
The subsequent processes in Valen handle these multi-viewpoint representation matrices in parallel and on an equal footing.
The method of processing RPM images into multi-viewpoint representations has been applied in previous works \cite{RS,SAVIR-T}. 

\begin{figure}[htp]\centering
	\includegraphics[trim=0cm 0cm 0cm 0cm, clip, width=8.5
 cm]{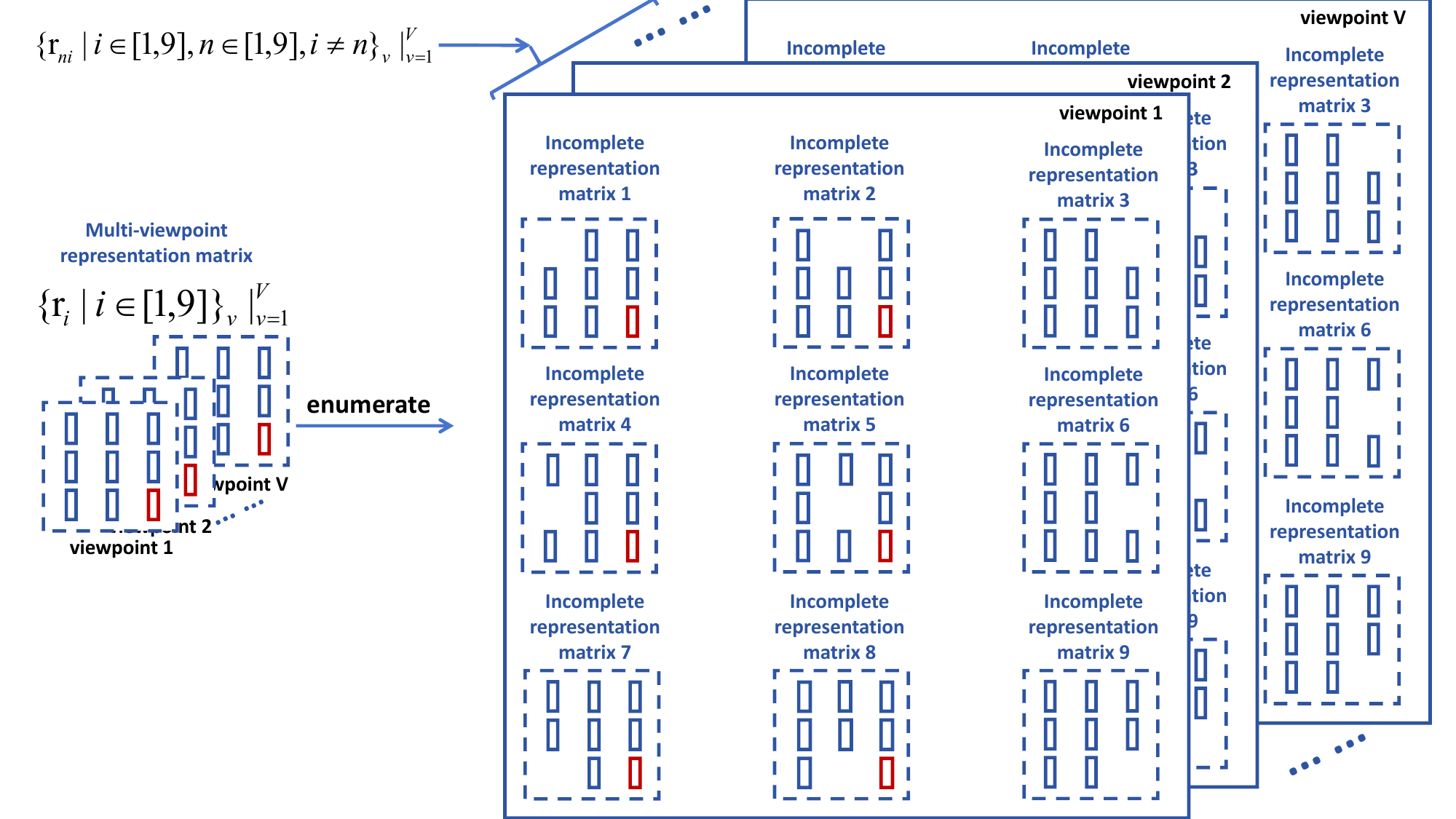}
	\caption{Enumeration results of incomplete matrix}
\label{Enumeration results of matrix}
\end{figure}

Subsequently, these multi-viewpoint representations $\{r_i | i \in [1,9]\}_v|_{v=1}^V$  will be enumerated by Valen to generate all possible incomplete representation matrices. The enumeration results are illustrated in Figure \ref{Enumeration results of matrix}.
We denote these enumerated multi-viewpoint incomplete representation matrices as $\{r_{ni}|n\in [1,9], i\in[1,9], i\neq n\}_v|_{v=1}^V$, where $n$ represents the index of the incomplete representation matrix, and $i$ still represents the index of the representation within the incomplete matrix. 
Notably, the last item of the enumeration is the matrix with an incomplete option position, denoted as $\{r_{9i}|i\in[1,8]\}_v|_{v=1}^V$. 

\subsubsection{Extraction of progressive patterns from incomplete matrices}
\begin{figure}[htp]\centering
	\includegraphics[trim=0cm 0cm 0cm 0cm, clip, width=8
 cm]{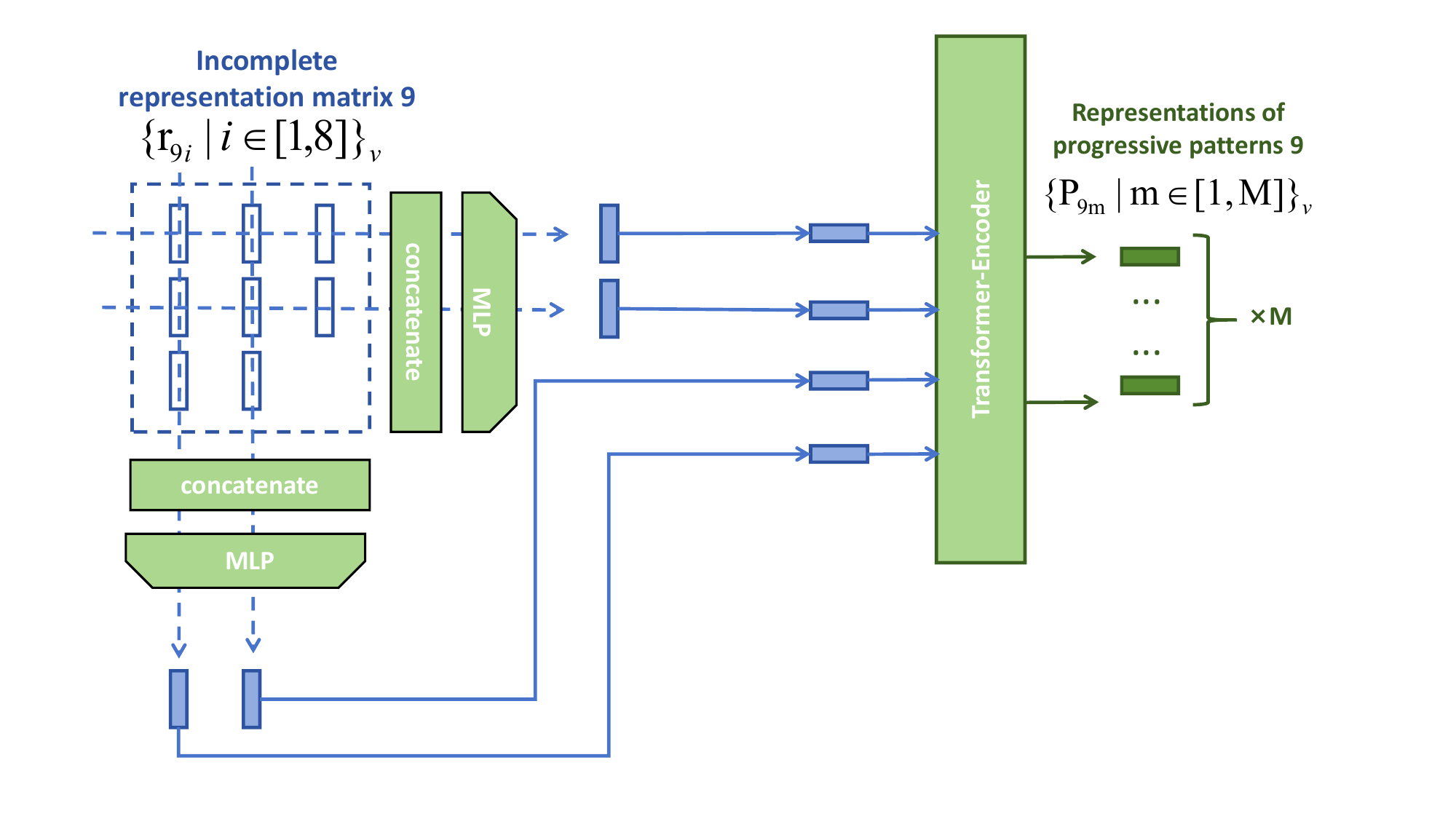}
	\caption{Extraction of progressive patterns from incomplete matrix}
\label{Extraction of progressive patterns from incomplete matrix}
\end{figure}

In Valen, this process is undertaken by a Transformer-Encoder. Specifically, we use a standard Transformer-Encoder to process all the previously enumerated multi-viewpoint incomplete representation matrices, with the processing procedure illustrated in Figure \ref{Extraction of progressive patterns from incomplete matrix}.
The figure illustrates the process of the Transformer-Encoder handling $\{r_{9i} | i \in [1,8]\}_v$, which represents the last item of the enumeration results of incomplete representation matrices from a specific viewpoint $v$. The process of handling other incomplete representation matrices $\{r_{ni} |n\in[1,8] ,i \in [1,9],i\neq n\}_v$ is similar to the one shown in the figure, with no essential differences.

In the Figure \ref{Extraction of progressive patterns from incomplete matrix}, we can observe that we use an MLP to extract all row and column progressive information from the incomplete representation matrix $\{r_{9i} | i \in [1,8]\}_v$. Subsequently, we process all four row and column progressive information using a standard Transformer-Encoder to obtain $M$ representations of progressive patterns. $M$ is a hyperparameter that can be freely set, and in this paper, it is set to 2. The impact of setting $M$ to other values on Valen is not explored in this paper. This asymmetric seq-to-seq implementation (four-to-$M$) has been introduced in related techniques of ViT\cite{ViT}. After processing all the
incomplete representation matrices $\{r_{ni} |n\in[1,9], i \in [1,9],i\neq n\}_v|_{v=1}^V$ using the Transformer-Encoder, we obtain all multi-viewpoint progressive pattern representations $\{P_{nm}|n\in[1,9],m \in[1,M]\}_v|_{v=1}^V$.

\subsubsection{Consistency analysis of progressive patterns
}
\begin{figure}[htp]\centering
	\includegraphics[trim=0cm 0cm 0cm 0cm, clip, width=8.5cm]{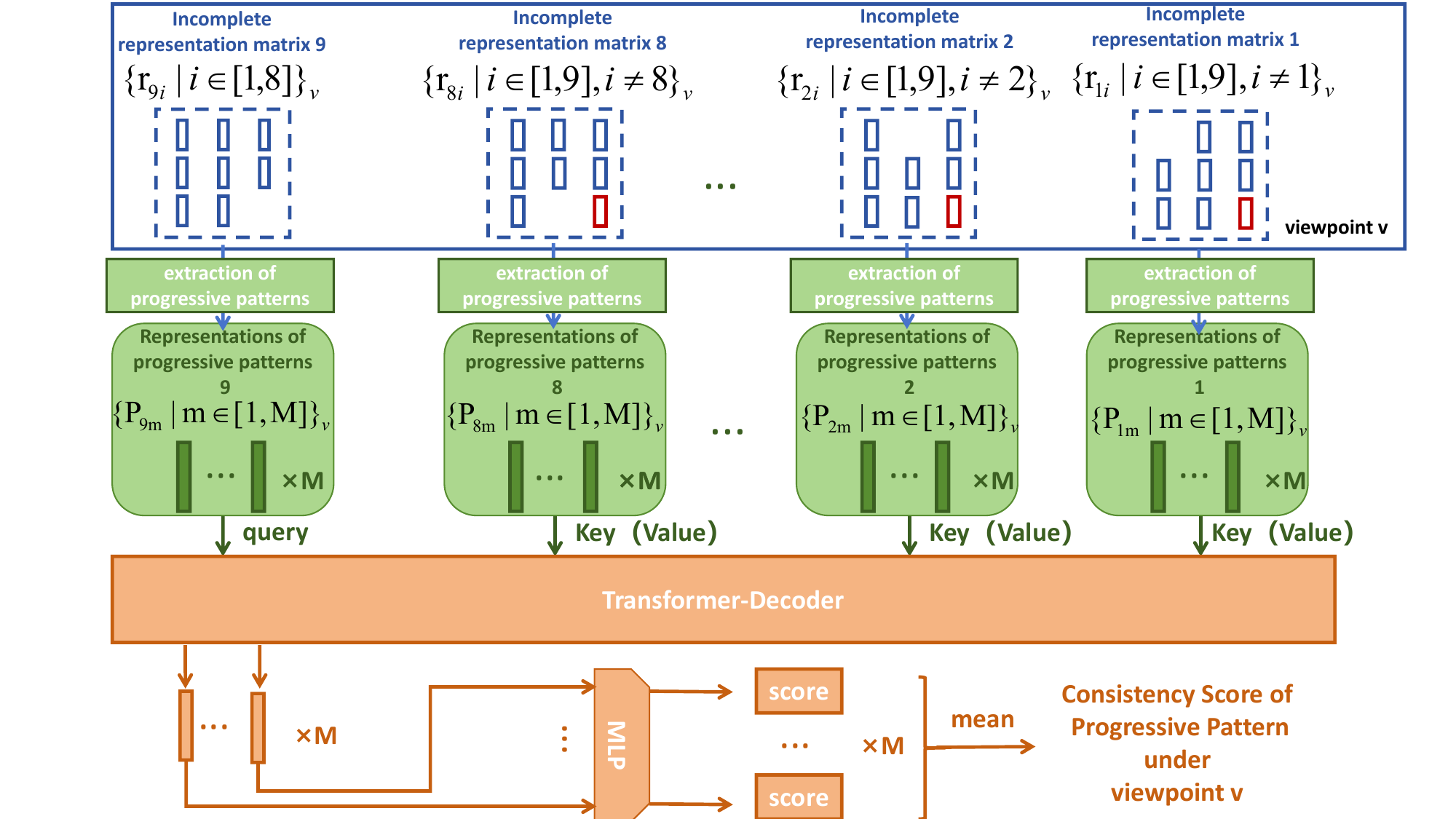}
	\caption{Evaluating the consistency of the progressive patterns representation $\{P_{nm}|n\in[1,9],m \in[1,M]\}_v$}
\label{evaluating the consistency of the progressive patterns representation}
\end{figure}

In this process, Valen evaluates the consistency among these multi-viewpoint progressive pattern representations $\{P_{nm}|n\in[1,9],m \in[1,M]\}_v|_{v=1}^V$. 
As illustrated in Figure \ref{evaluating the consistency of the progressive patterns representation}, Valen conducts this consistency evaluation among the representations $\{P_{nm}|n\in[1,9],m \in[1,M]\}_v$ from each specific viewpoint $v$ using a Transformer-Decoder.

In the Figure \ref{evaluating the consistency of the progressive patterns representation}, the progressive pattern representations $\{P_{9m}|m \in[1,M]\}_v$ extracted from the incomplete representation matrix $\{r_{9i} | i \in [1,8]\}_v$ are used as queries, while the remaining progressive pattern representations $\{P_{nm} | n\in [1,8], m \in [1,M]\}_v$, which are extracted from the incomplete representation matrices $\{r_{ni} | n\in [1,8], i \in [1,9], i\neq n\}_v$, act as key-value pairs for the Transformer-Decoder to compute the output vector set.
The $M$ output vectors are then individually mapped to scores by an MLP and summed to yield the consistency score of the progressive patterns under viewpoint $v$.

Subsequently, we compute the average of the progressive pattern consistency scores across all viewpoints to obtain the overall viewpoint score for the corresponding option. Finally, cross-entropy loss is employed to align the score distribution with the ground-truth option, thereby optimizing all learnable parameters involved in the aforementioned three processes.

\subsection{Valen targeting Bongard-Logo problem}

We regard Valen as a strong baseline for image progression pattern reasoning and as a model capable of tackling image clustering reasoning problem such as Bongard-Logo. To adapt Valen to Bongard-Logo, we transform the task accordingly.

In a given Bongard-Logo instance, we denote the 14 images as $\{x_i | i\in[1,14]\}$. Specifically, the images in the primary group are $\{x_i | i\in[1,6]\}$, while the images in the auxiliary group are $\{x_i | i\in[8,13]\}$. Furthermore, $x_7$ is to be assigned to the primary group, and $x_{14}$ is to be assigned to the auxiliary group.

\begin{figure}[htp]\centering
	\includegraphics[trim=0cm 0cm 0cm 0cm, clip, width=8.5
 cm]{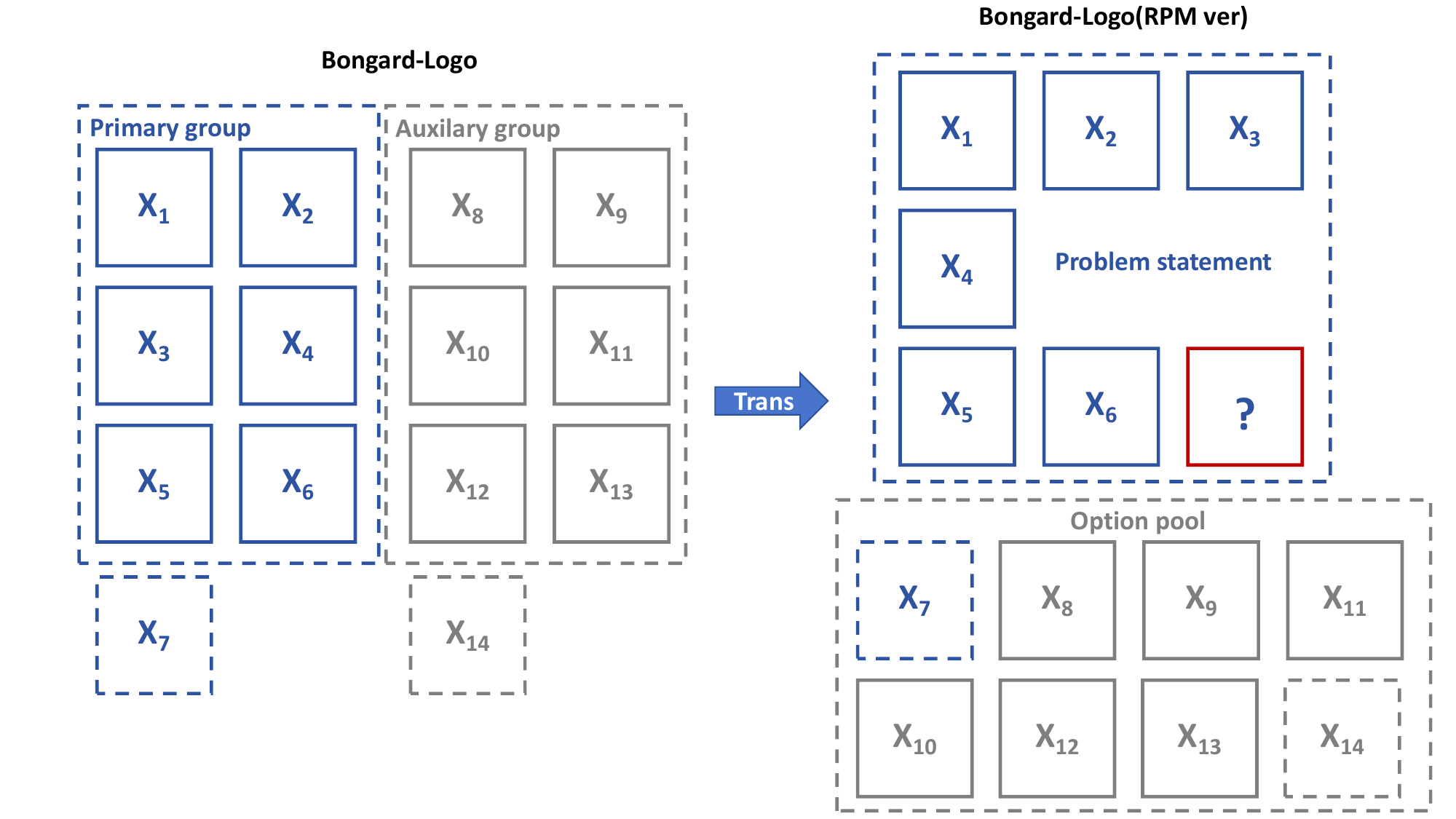}
	\caption{Transformation of Bongard-Logo problem}
\label{Transformation of Bongard-Logo problem}
\end{figure}

This paper proposes that image clustering reasoning problems, such as Bongard-Logo, can be approached from different perspectives. Specifically, these tasks can be recast as option-evaluation problems, analogous to those in the RPM task. To elaborate, the images in the primary group $\{x_i|i\in [1,6]\}$ can be regarded as a problem statement akin to that in RPM, while the test images and those in the auxiliary group $\{x_i|i\in [7,14]\}$ act as an option pool analogous to the one in RPM. This transformation is shown in Figure \ref{Transformation of Bongard-Logo problem}.
After this transformation, Valen can be effectively applied to image clustering reasoning problems such as Bongard-Logo.

It is noteworthy that when transferring Valen to Bongard-Logo, some adjustments are still necessary to accommodate the data characteristics of Bongard-Logo. Specifically, due to the scarcity of training instances in the Bongard-Logo problem, the ViT-based image representation extractor may struggle to perform effectively \cite{ViT,RS}. Furthermore, although we transform the Bongard-Logo problem into an RPM-like format, it is unnecessary to analyze the progressive patterns among Bongard-Logo images, as such patterns do not exist in clustering reasoning problems like Bongard-Logo.

\begin{figure}[htp]\centering
	\includegraphics[trim=0cm 8cm 0cm 0cm, clip, width=8.5
 cm]{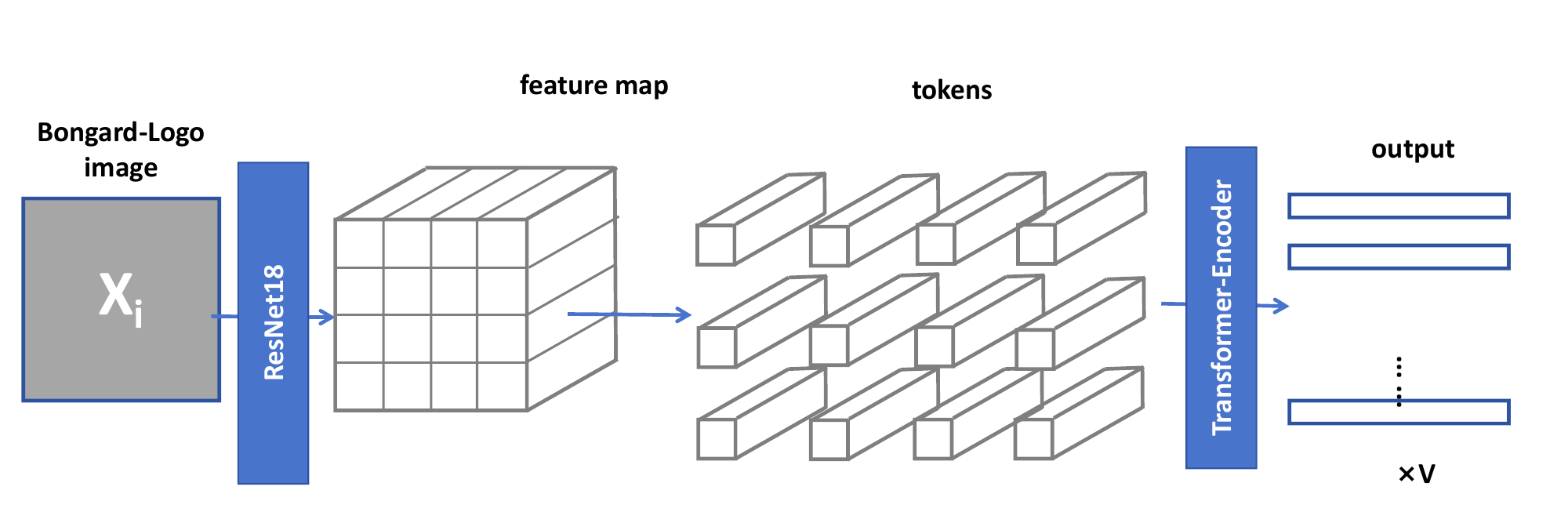}
	\caption{Representation extraction of Bongard-Logo image}
\label{Representation extraction of Bongard-Logo image}
\end{figure}

Therefore, we have redesigned the image representation extractor, originally illustrated in Figure \ref{representation extraction of matrix}, into a convolutional network concatenated with a Transformer-Encoder. The new design is illustrated in Figure \ref{Representation extraction of Bongard-Logo image}. 
In this figure, when a Bongard-Logo image is processed by Valen's new feature extractor, it is first encoded into a feature map by ResNet18. This map is then decomposed into multiple tokens and fed into a Transformer-Encoder to obtain multi-viewpoint representations. The design combines CNN’s sample efficiency with ViT’s multi-viewpoint modeling capacity \cite{ImageNet,ResNet,ViT,RS,SAVIR-T} and has been used in previous work \cite{SAVIR-T}.

\begin{figure}[htp]\centering
	\includegraphics[trim=0cm 7cm 5cm 0cm, clip, width=8.5
 cm]{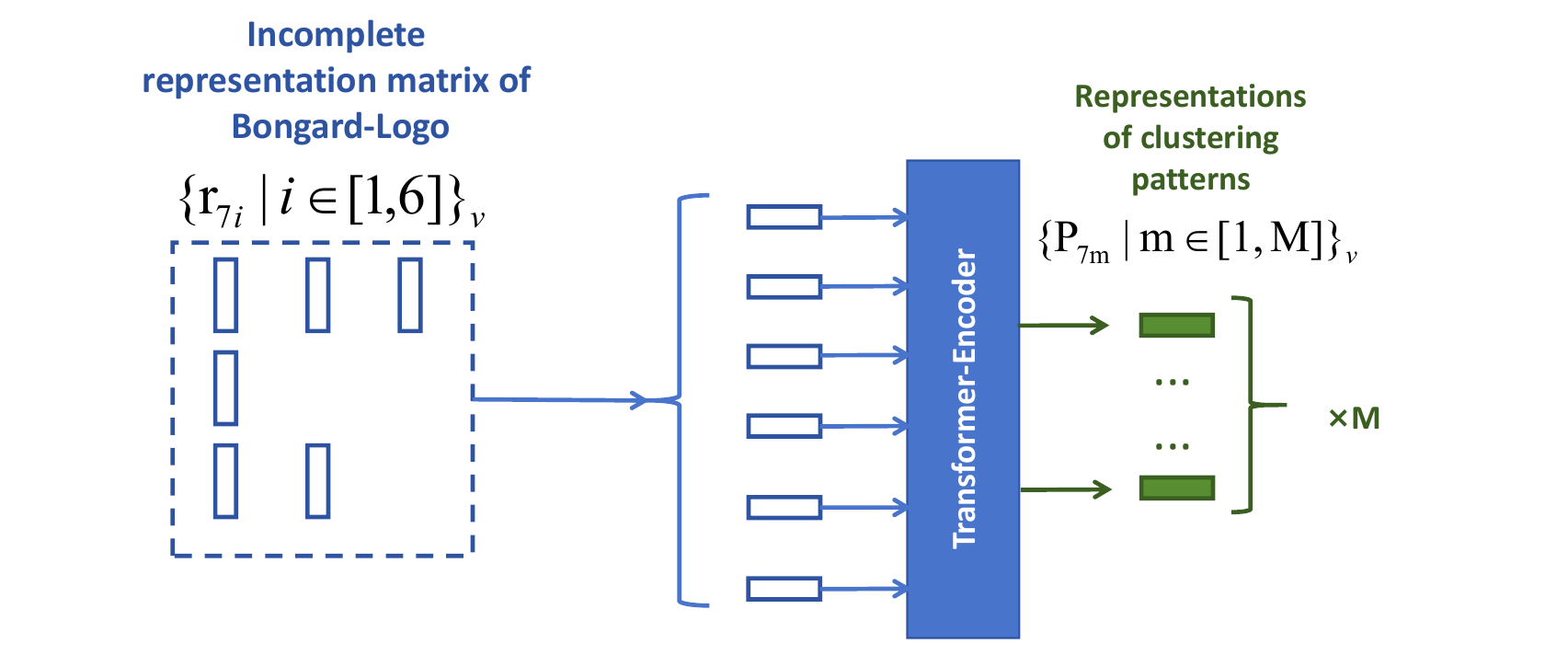}
	\caption{Pattern extraction of Bongard-Logo incomplete matrix}
\label{Pattern extraction of Bongard-Logo incomplete matrix}
\end{figure}

Additionally, we have simplified the incomplete matrix progressive pattern extractor, which was originally intended to be structured as shown in Figure \ref{Extraction of progressive patterns from incomplete matrix}, into the form illustrated in Figure \ref{Pattern extraction of Bongard-Logo incomplete matrix}.
In this figure, we abandon the use of MLP to extract the row and column progressive information from the incomplete matrix in Bongard-Logo, because the transformed Bongard-Logo problem only contains clustering patterns and not progressive patterns. We then feed the representations of all images within the Bongard-Logo incomplete matrix as tokens into the Transformer-Encoder to extract $M$ clustering patterns. When processing Bongard-Logo, $M$ is set to 6.

\section{Tine, a Method for Advancing the Solution Distribution Boundary}

Current end-to-end visual abstract reasoning solvers, including Valen, are keen on assigning probabilities to each option. Although their loss functions are diverse and their backbone architectures vary significantly, most adhere to this paradigm. We term such solvers ``probability-highlighting solvers".

Therefore, we can interpret these solvers as a conditional probability distribution  governing the solutions to abstract reasoning problems. Taking RPM solvers as examples, they can be abstracted as a conditional distribution, denoted as $P(x_\alpha | \{x_i\}_{i=1}^8;\theta )$. (Accordingly, when Valen is applied to Bongard-Logo problems, it can be abstracted as $P(x_\alpha | \{x_i\}_{i=1}^6;\theta )$). 
Here, $\{x_i\}_{i=1}^8$ is the statement of the reasoning problem, $\theta$ represents the learnable parameters of the solver, and $x_\alpha$ refers to the option that is pending verification.
This distribution gives the probability that the solver's parameters $\theta$ deem $x_\alpha$ to be the solution to the context $\{x_i\}_{i=1}^8$.

The optimization of a probability-highlighting solver corresponds to tuning the parameters $\theta$ within the conditional distribution $P(x_\alpha | \{x_i\}_{i=1}^8;\theta )$. Although a wide range of loss functions can be employed, they all share the same objective: adjust parameters $\theta$ so that the distribution $P(x_\alpha | \{x_i\}_{i=1}^8;\theta )$ assigns high probability to the primary samples and low probability to the auxiliary samples.
Such an optimization objective makes us realize that the optimization process of the probability-highlighting solver is not aimed at fitting the distribution of correct solutions to visual abstract reasoning problems, but rather is dedicated to fitting a distribution jointly delimited by primary and auxiliary samples.

To simplify the explanation of our findings, we assume that the conditional probability distribution $P(x_\alpha | \{x_i\}_{i=1}^8;\theta )$ follows a Gaussian distribution, so we write it explicitly as $P(x_\alpha | \{x_i\}_{i=1}^8;\mu_\theta, \sigma_\theta )$, where $\mu_\theta$, $\sigma_\theta$ represent the mean and standard deviation, respectively. Then, the negative logarithm of this Gaussian density can be expressed as follows:
\begin{align}\label{P-RPMSoLvers}
    -\log(P(x_\alpha|\{x_i\}_{i=1}^8;\mu_\theta, \sigma_\theta )) = \frac{{(x_{\alpha} - \mu_\theta)^{2}}}{{2\sigma^{2}_\theta}} - \log(2\pi|\,\sigma_\theta|\,)
\end{align}
Accordingly, the optimization objective is to tune the values of $\mu_\theta$ and $\sigma_\theta $ so that when primary samples (i.e., the correct RPM options or samples within the primary group in Bongard-Logo) are plugged into formula (\ref{P-RPMSoLvers}), the formula returns a smaller value; conversely, when auxiliary samples (i.e., the incorrect RPM options or the samples within the auxiliary group in a Bongard-Logo problem) are plugged into formula (\ref{P-RPMSoLvers}), the formula returns a larger value.

Evidently, minimizing the return value of the formula (\ref{P-RPMSoLvers}) amounts to minimizing the Euclidean distance between $\mu_\theta$ and $x_\alpha$; maximizing it corresponds to minimizing the value of $\sigma_\theta$.
In other words, the probability-highlighting solver positions the mean of the solution distribution $P(x_\alpha | \{x_i\}_{i=1}^8;\theta )$ for the reasoning problem through primary samples, while the standard deviation of the solution distribution is constrained by auxiliary samples.

It appears to be a discouraging reality that the learning objective of a probability-highlighting solver is not to determine the distribution of the correct solutions to reasoning problems, but rather to approximate a distribution that is delimited by both primary and auxiliary samples within those problems.
This suggests that the reasoning accuracy of the probability-highlighting solver is constrained. The standard deviation of the distribution delimited by the auxiliary samples is necessarily greater than or equal to the standard deviation of the correct solution distribution. In other words, the distribution delimited by both primary and auxiliary samples is often much coarser than the distribution of correct solutions. Aiming to approximate the former as the optimization objective is akin to ``aiming for mediocrity and achieving even less".  This discouraging reality can be visualized in Figure \ref{Schematic diagram of the relative relationship between the solution distribution, the correct solution distribution, and the primary and auxiliary samples.}. Note that this figure is intended to schematically represent our analysis results, rather than to depict the actual form of the solution distribution or the correct solution distribution.
\begin{figure}[htp]\centering
	\includegraphics[trim=0cm 3cm 2cm 0cm, clip, width=7.3
 cm]{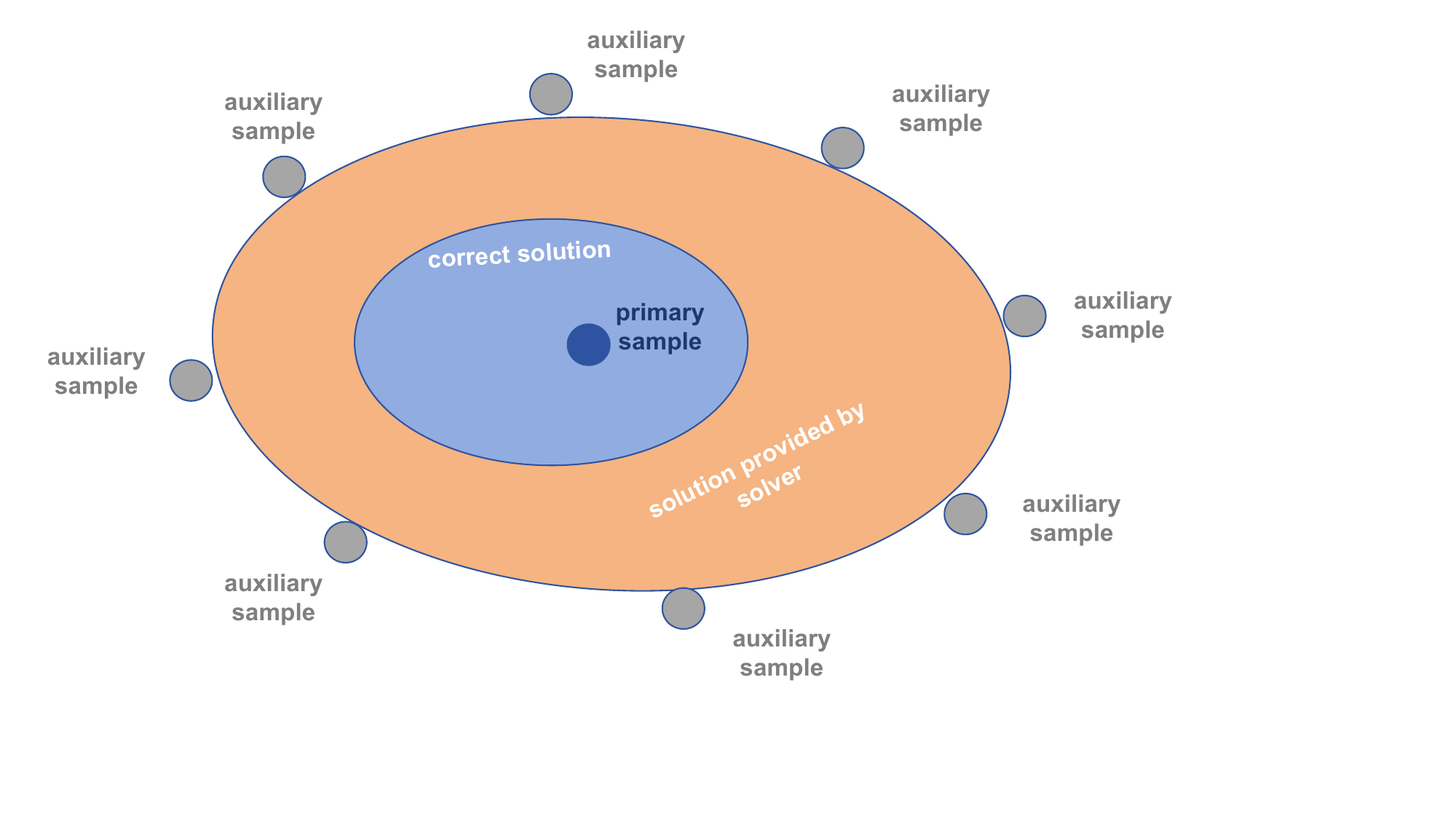}
	\caption{Schematic diagram of the relative relationship between the solution distribution, the correct solution distribution, and the primary and auxiliary samples.}
\label{Schematic diagram of the relative relationship between the solution distribution, the correct solution distribution, and the primary and auxiliary samples.}
\end{figure}

\begin{figure}[htp]\centering
	\includegraphics[trim=0cm 3cm 2cm 0cm, clip, width=8.5
 cm]{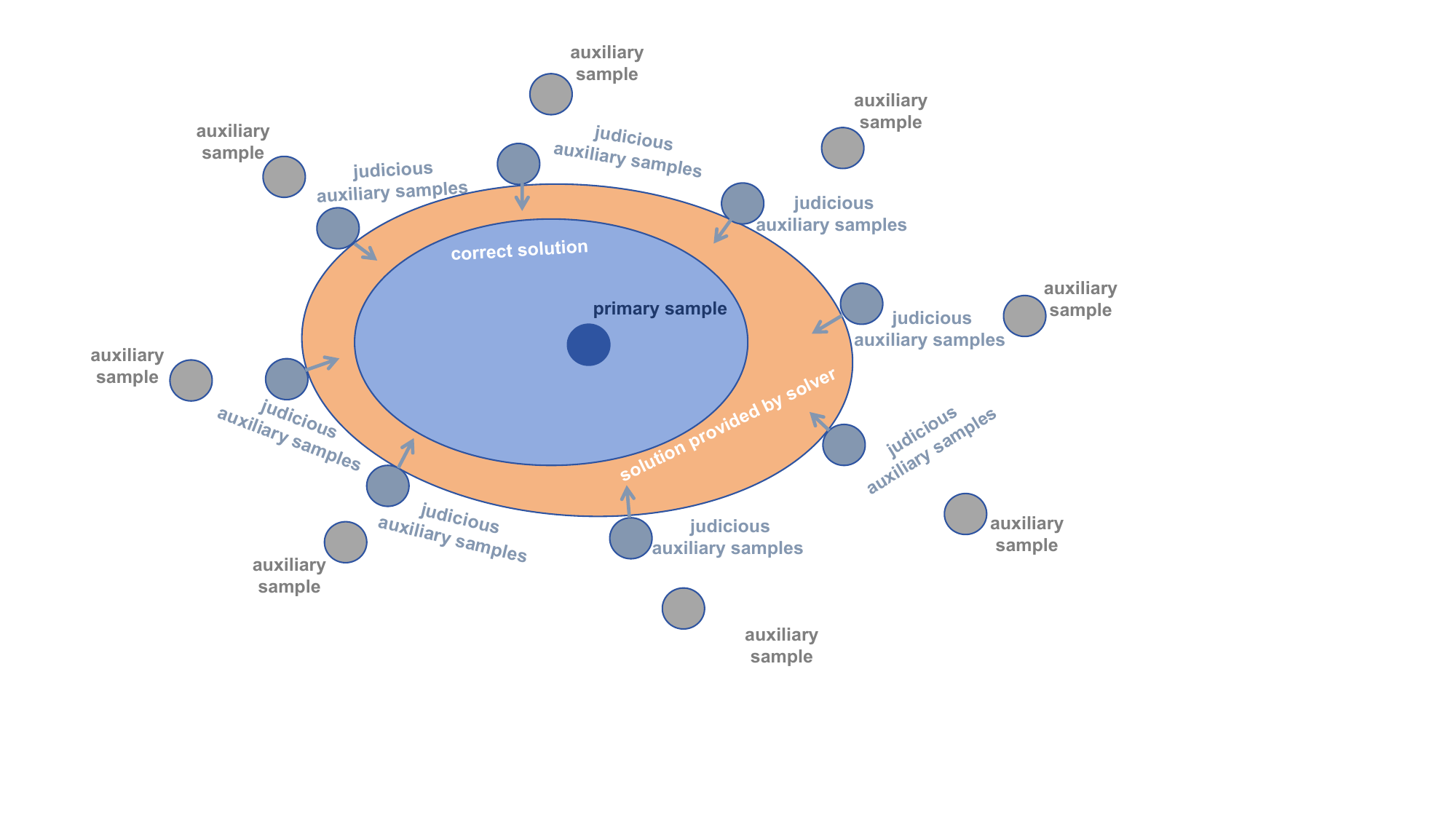}
	\caption{{J}udicious auxiliary samples}
\label{judicious auxiliary samples}
\end{figure}

This paper argues that to further improve the reasoning accuracy of Valen in solving RPM problems and Bongard-Logo problems, it is necessary to provide Valen with more judicious auxiliary samples. By ``more judicious auxiliary samples'', we refer to those that can better delimit the solution distribution provided by Valen.
These more judicious auxiliary samples have the potential to narrow down the standard deviation of the solution distribution. The visualization of these more judicious auxiliary samples can be found in Figure \ref{judicious auxiliary samples}.
As shown in this figure, the judicious auxiliary samples we desire appear more authentic or closer to primary samples from Valen's perspective than the original auxiliary samples.
However, since judiciousness is evaluated from Valen's perspective, it lacks human interpretability; samples with smaller or fewer attribute shifts are not necessarily deemed more judicious by Valen.

To this end, this paper argues that we can use adversarial learning to supply Valen with auxiliary samples it deems more judicious. Given this insight, we design the Tine (Technique for edge Identification of solution distribution based on adversarial Networks) method. Specifically, we adopt adversarial generation techniques similar to GAN \cite{GAN} or WGAN \cite{WGAN} to generate more judicious samples for Valen. However, unlike GAN or WGAN, whose objective is to train robust and high-performance generators, we use adversarial learning to refine Valen, which acts as the discriminator.

Our objective is to design an auxiliary sample generator for Valen. Its task is to produce samples that receive higher probabilities from Valen than the original auxiliary samples. Accordingly, Valen's role is to classify these generated samples as low-scoring ones. During training, the parameters of Valen and the generator are updated alternately and exclusively, undergoing standard adversarial refinement.

\begin{figure}[htp]\centering
	\includegraphics[trim=0cm 0cm 0cm 0cm, clip, width=8.5
 cm]{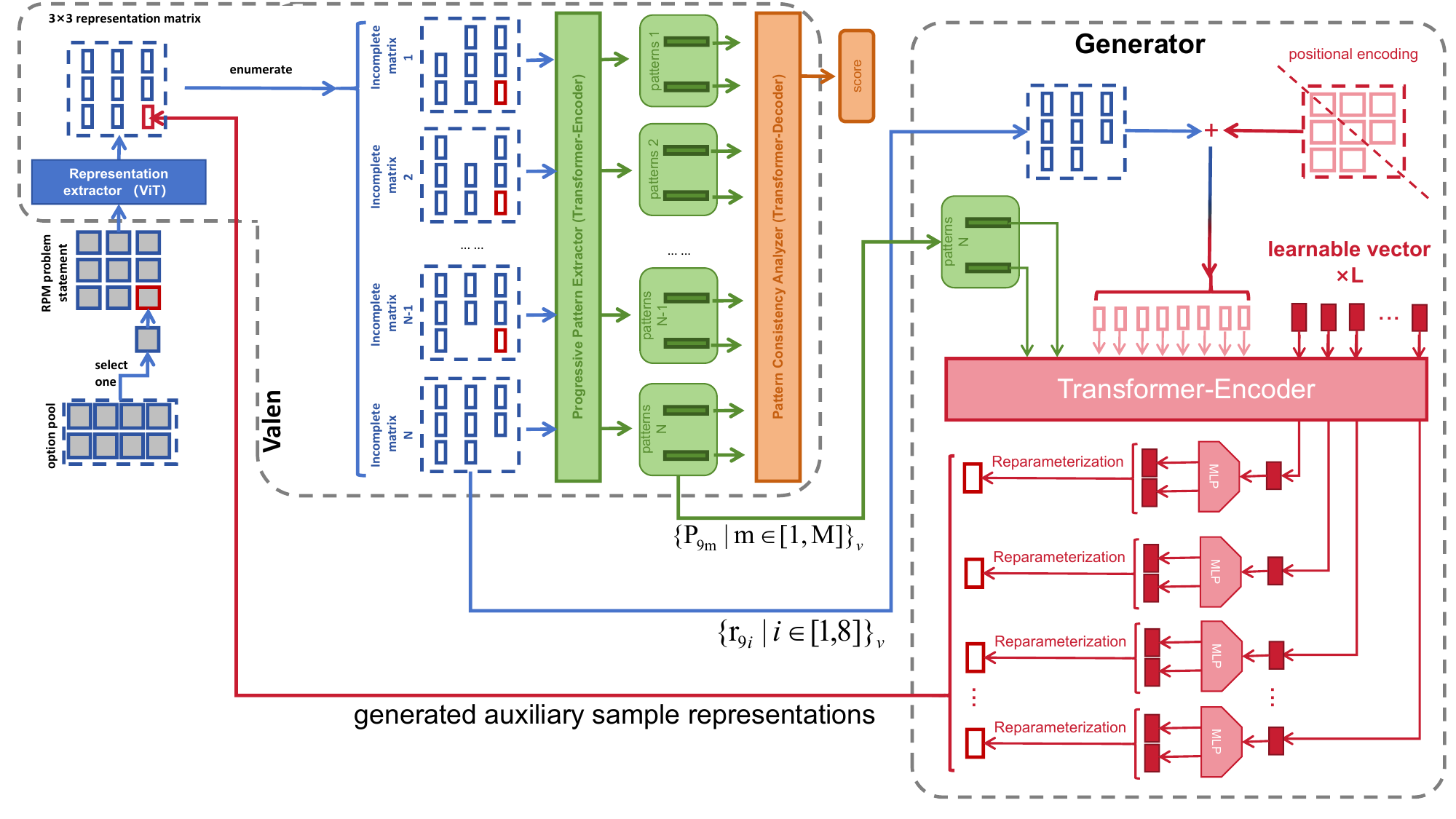}
	\caption{The structure of the generator associated with Valen}
\label{The structure of the generator associated with Valen.}
\end{figure}

Given that our goal is to generate more judicious auxiliary samples from Valen's perspective, we are not concerned with their interpretability, attribute values, or whether they strictly conform to the nature of reasoning problems. Consequently, rather than forcing the generator to produce pixel-level auxiliary samples, which would incur greater computational overhead and optimization difficulties, we opt to generate representations of judicious auxiliary samples. By outputting only multi-viewpoint representations, the generator enjoys a simpler architecture and becomes easier to optimize. The generator is shown in Figure \ref{The structure of the generator associated with Valen.}. Since we have already transformed the Bongard-Logo problem into an RPM-like problem, we will not further elaborate on the structure of the generator for the Bongard-Logo version.

\begin{figure}[htp]\centering
	\includegraphics[trim=0cm 0cm 0cm 0cm, clip, width=8.5
 cm]{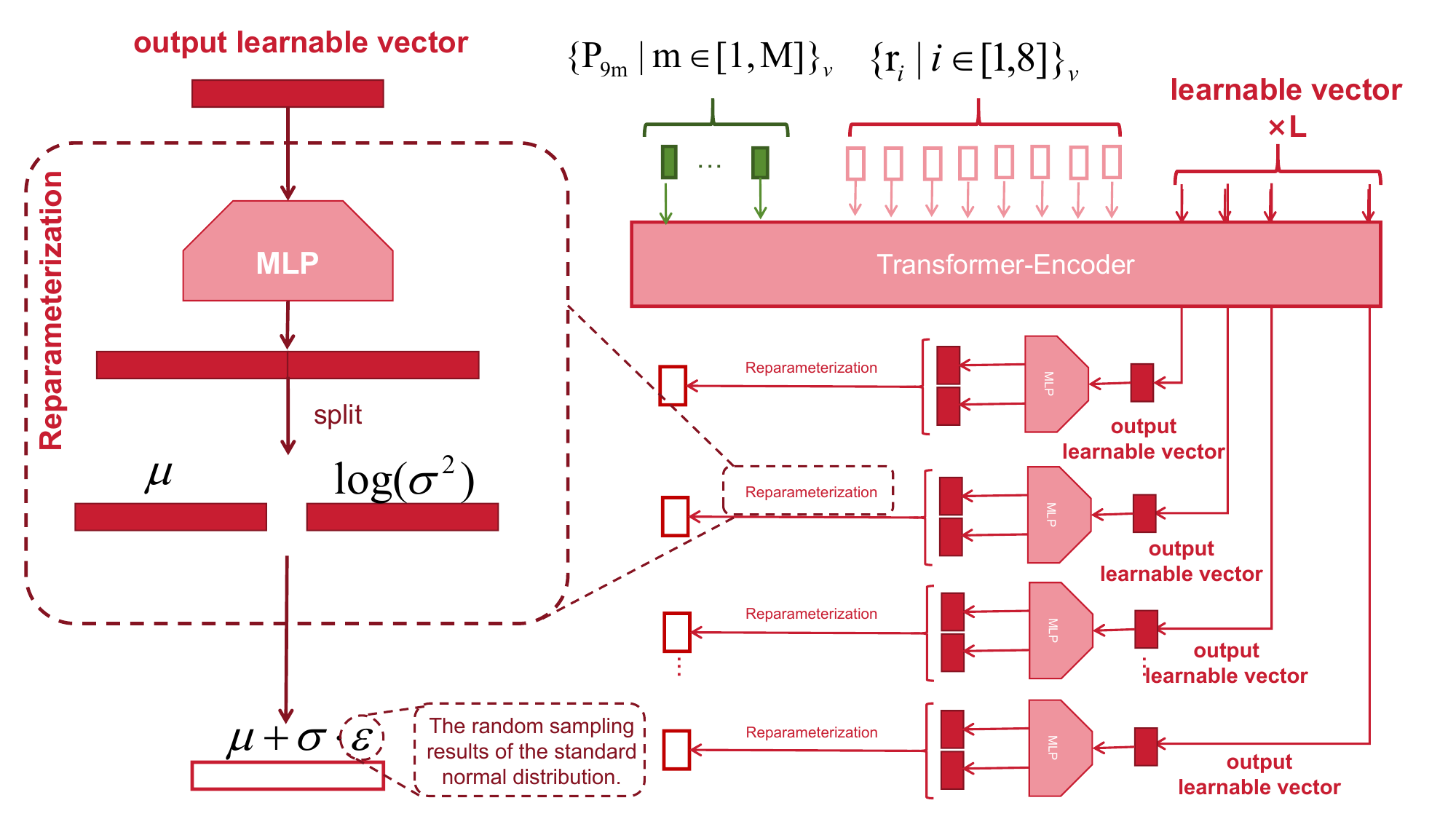}
	\caption{Reparameterization}
\label{reparameterization}
\end{figure}

As depicted in Figure \ref{The structure of the generator associated with Valen.}, the left gray dashed box encloses the Valen model, while the right gray dashed box encloses the generator designed for Valen. The backbone of this generator is a standard Transformer-Encoder, which generates representations of more judicious auxiliary samples based on the representations of the abstract reasoning problem statement $\{r_i | i \in [1,8]\}_v$, their progressive patterns $\{P_{9m}|m \in[1,M]\}_v$, and $L$ learnable vectors. Both the representations and the progressive patterns of the statement are provided by Valen, while the $L$  learnable vectors must be additionally introduced for the generator. It is noteworthy that when dealing with PGM problems within RPM tasks, we have designed special positional encoding for the representations of the problem statement. This positional encoding can be specifically attached to incomplete matrices with missing option positions and is set symmetrically along the diagonal. 
After the Transformer-Encoder processes the input representations (i.e., $\{r_i | i \in [1,8]\}_v$, $\{P_{9m}|m \in[1,M]\}_v$ and $L$ learnable vectors) and generates the output, we extract and save those $L$ learnable vectors.
These processed vectors then pass through an MLP that doubles their dimension.
The resulting $L$ doubled-length vectors are transformed back into $L$ vectors using the reparameterization technique \cite{VAE}. The details of this technique are shown in Figure \ref{reparameterization}.
These $L$ reparameterized vectors are regarded as the representations of $L$ auxiliary samples. It is worth noting that $L$ is a hyperparameter that can be freely set, determining how many judicious auxiliary samples Tine generates for Valen in one instance. Here, $L$ is set to 9.

In short, our Tine method was conceived after summarizing the working principles of probability-highlighting solvers. We contend that such solvers establish a solution distribution for every instance of a reasoning problem and adjust it so that primary samples receive high probability while auxiliary samples receive low probability. This means their optimization target is not to align with the distribution of correct solutions to the reasoning problems, but rather to align with the distribution jointly delimited by primary and auxiliary samples. Unfortunately, this delimited distribution is often much coarser than the correct one. To bring the former closer to the latter, we designed Tine based on adversarial learning.

\section{Funny: Method for Planning the Distribution of Reasoning Problem Samples.}

Based on the analysis in the previous section, auxiliary samples provided in reasoning tasks constrain the standard deviation of the solution distribution $P(x_\alpha | \{x_i\}_{i=1}^8;\theta )$, while primary samples locate the mean of the solution distribution $P(x_\alpha | \{x_i\}_{i=1}^8;\theta )$. The Tine method generates more judicious auxiliary samples to further constrain the standard deviation. However, it remains unclear whether this constraint is insufficient or excessive. Moreover, being rooted in adversarial learning, Tine inevitably inherits the associated drawbacks, such as unstable training and difficulty in assessing convergence. Faced with these challenges, we propose the Funny (Framework Utilizing Neural Networks for Yielding Image Representation Distribution) method.

Our Funny method posits that, rather than laboriously adopting adversarial learning to make the solver's distribution $P(x_\alpha | \{x_i\}_{i=1}^8;\theta )$ close to the correct solution distribution, it is more effective to directly specify the range of correct solutions or to explicitly characterize the correct solution distribution. Toward this goal, the present section dissects the correct solutions for both image progression pattern and image clustering reasoning problems.

\subsection{Key attributes and non-key attributes}

According to the design of RPM problems, the attributes of an image are divided into key attributes that are relevant to reasoning and non-key attributes that serve as distractors. The latter are introduced solely to interfere with the solver and to evaluate its ability to filter irrelevant information. It is critical to note that attributes alone determine the identity of an RPM image: any two images that share identical key attributes have the same identity in the reasoning process, regardless of their non-key differences. Consequently, any sample whose key attributes satisfy the governing pattern is deemed a correct solution. 
For example, in the RAVEN problem\cite{RAVENdataset} the rotation angle of an entity is treated as a non-key attribute, whereas color, size, shape, and position are key; thus, a white, medium-sized hexagon positioned in the center retains its reasoning identity no matter how it is rotated and cannot be transformed by rotation from a correct solution into an incorrect one, or vice versa.

The design of correct solutions for Bongard-Logo problems is much more complex.
The reasoning identity of a Bongard-Logo sample is determined by its instance context, which also dictates the key and non-key attributes. Some instances focus on macroscopic attributes such as closure, concavity and convexity, while others concentrate on detailed attributes such as the number of entities and their spatial order. This contrasts with RPM problems, where key and non-key attributes are consistent across all instances.

Due to the differences between Bongard-Logo and RPM problems in determining correct solutions and in distinguishing key from non-key attributes, we apply tailored variants of our Funny approach to the two problem types; each will be discussed separately.

\subsection{Funny for RPM problems}
\begin{figure}[htp]\centering
	\includegraphics[trim=0cm 6cm 0cm 0cm, clip, width=8.5   
 cm]{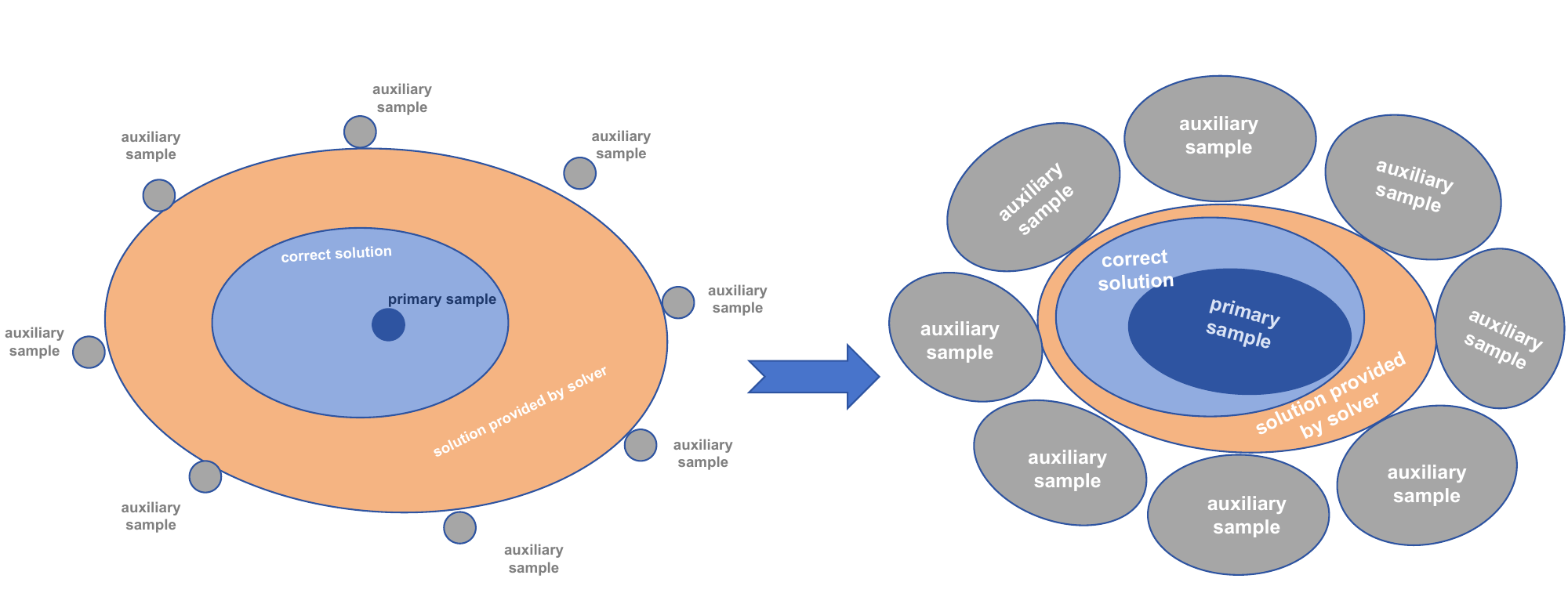}
	\caption{Modeling the representation distribution as a mixture of Gaussian distributions.}
\label{Modeling the representation distribution as a mixture of Gaussian distributions.}
\end{figure}

Given that the key attributes and non-key attributes in the RPM problem are fixed, with key attributes being key across all instances and non-key attributes being unfavored in any instance, this paper proposes that the representations of samples with the same key attributes can be mapped into a single Gaussian distribution. Within this Gaussian distribution, the key attributes of the samples are identical, but the non-key attributes vary. Consequently, the distribution of representations for all samples in the entire RPM problem is formulated as a mixture of Gaussian distributions \cite{GMM}. By exposing Valen to this mixture instead of individual sample representations, we effectively clarify the distribution of correct solutions for Valen. Our intention is to achieve the transformation of the sample representation space as illustrated in Figure \ref{Modeling the representation distribution as a mixture of Gaussian distributions.}.
Therefore, the Funny method aims to plan the RPM sample distribution as a Gaussian mixture in which the key attributes define the component means.

\begin{figure}[htp]\centering
	\includegraphics[trim=5cm 4.2cm 0cm 6cm, clip, width=8.5   
 cm]{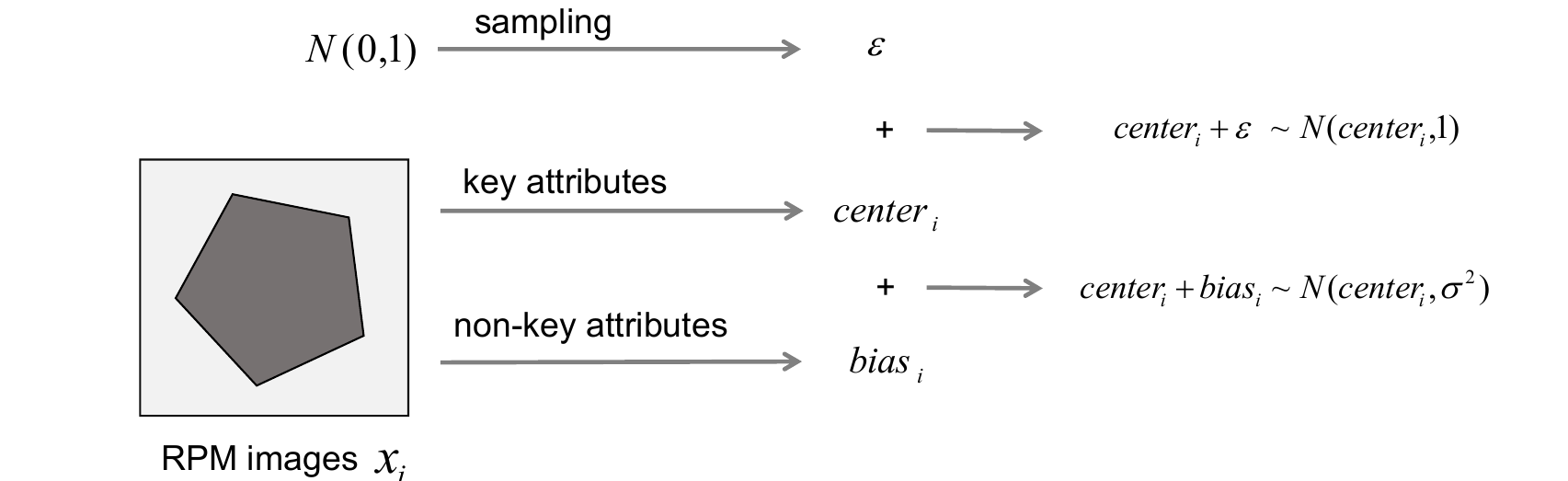}
	\caption{{The form of image attribute extraction adopted by Funny.} }
\label{The form of image attribute extraction adopted by Funny.}
\end{figure}

The design of the Funny method for tackling RPM problems is straightforward: Funny adopts the image attribute extraction scheme illustrated in Figure \ref{The form of image attribute extraction adopted by Funny.}.
In this figure, the Funny method extracts the key attributes and non-key attributes of RPM images, treating the key attributes as the center and the non-key attributes as biases relative to this center. Accordingly, we refer to the representation of the key attributes extracted from the RPM image $x_i$ as $center_i$, and that of the non-key attributes as $bias_i$.
Interestingly, the sum of $center_i$ and $bias_i$ can be viewed as a sample drawn from the Gaussian distribution $N(center_i, \sigma^2 )$, which represents a distribution with a mean of $center_i$ and an unknown variance $\sigma^2$. When the $bias_i$ follow a standard normal distribution, this Gaussian distribution can be considered as $N(center_i,1 )$, which is a distribution with a mean of $center_i$ and a variance of 1.
In this scenario, representations of samples with the same key attributes but different non-key attributes only differ in their biases, implying that the Funny method successfully maps representations of samples with the same key attributes into the same Gaussian distribution.

\begin{figure}[htp]\centering
	\includegraphics[trim=0cm 0cm 0cm 0cm, clip, width=8.5
 cm]{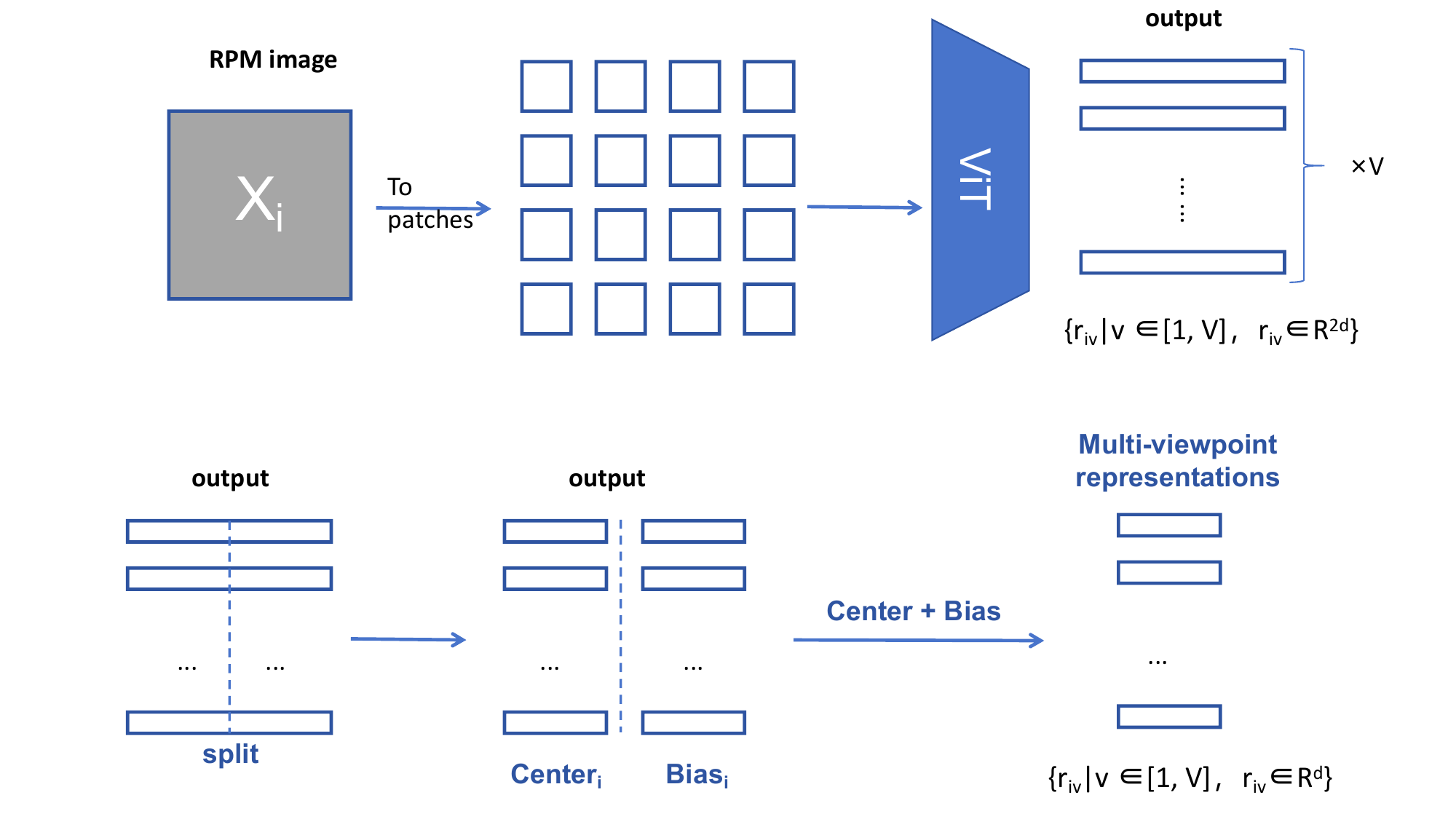}
	\caption{Funny representation extractor}
\label{Funny representation extractor}
\end{figure}

Based on this hypothesis, we design a multi-viewpoint image representation extractor as illustrated in Figure \ref{Funny representation extractor}, which replaces Valen's original representation extractor when equipped with the Funny method. 
As can be seen in this figure, Funny encodes the reasoning image $x_i$ into two sets of homomorphic multi-viewpoint representations, termed $center_i$ and $bias_i$ respectively. The final multi-viewpoint representation $r_i$ is obtained by element-wise summation of $center_i$ and $bias_i$.

However, the new representation extractor of the Funny method, which encodes images $x_i$ into $center_i$ and $bias_i$, still faces two challenges: (1) How to ensure that the $center_i$ stores key attributes while the $bias_i$ captures non-key attributes? (2) Given that the union of key and non-key attributes constitutes all attributes, how to guarantee that the combination of $center_i$ and $bias_i$ can store all attributes of the image $x_i$? These questions pose challenges to both the center and bias individually and collectively.

This paper argues that by enforcing $bias_i$ to follow a standard normal distribution $N(0,1)$ and simultaneously enforcing all random sampling results from a Gaussian distribution $N(center_i,1)$ to assume the same reasoning identity, it can be ensured that $center_i$ stores at least the key attributes. Under this constraint, $bias_i$, which follows a standard normal distribution $N(0,1)$, ceases to influence the reasoning, and $center_i$ takes full responsibility for the reasoning task, indicating that $center_i$ stores at least the key attributes at this point.

\begin{figure}[htp]\centering
	\includegraphics[trim=0cm 0cm 0cm 0cm, clip, width=8.5
 cm]{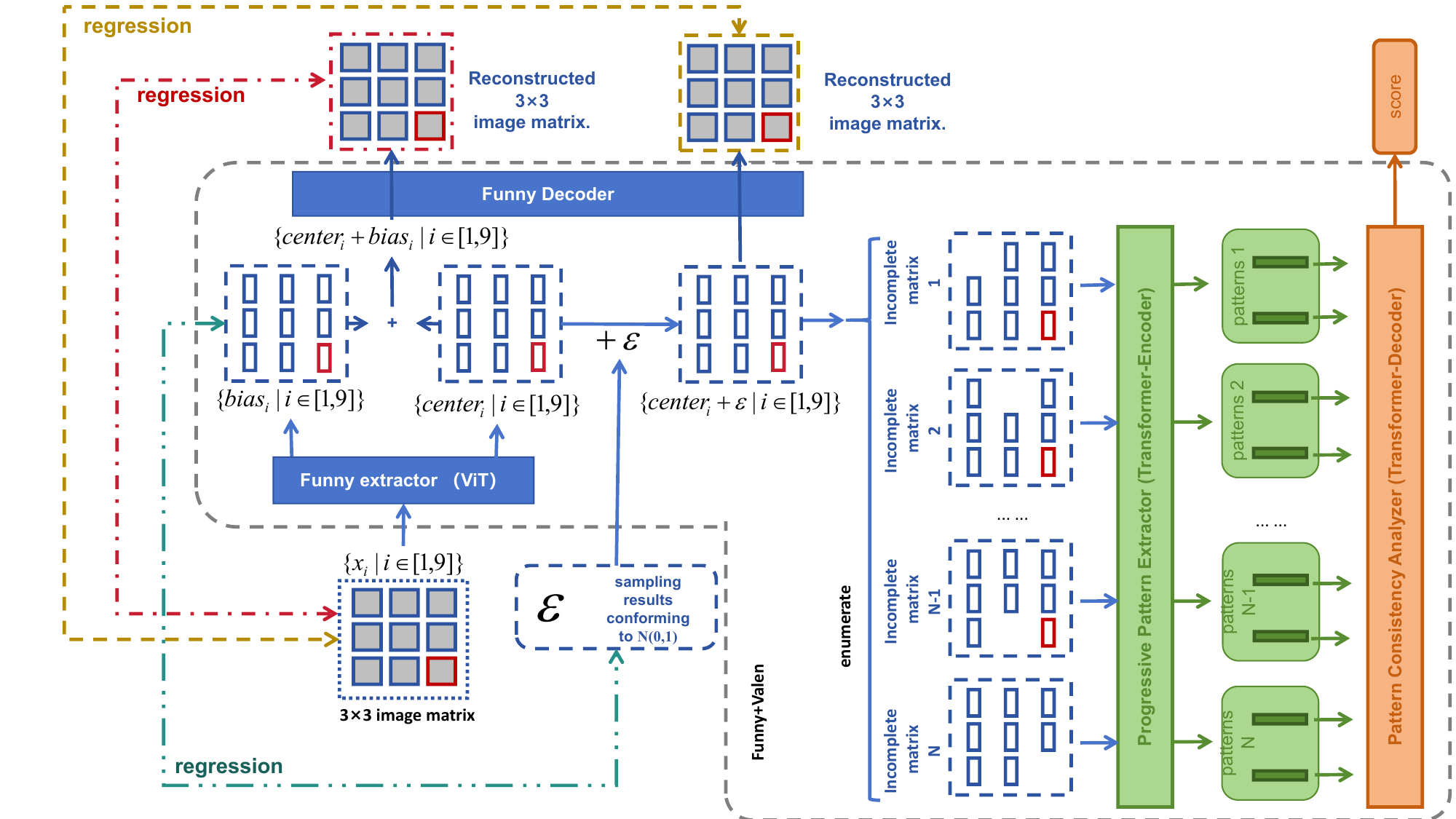}
	\caption{The Valen framework incorporates the Funny method}
\label{The Valen framework incorporates the Funny method}
\end{figure}

Furthermore, we propose that maximizing the mutual information between $center_i + bias_i$ and the image $x_i$ suffices to ensure that this combined representation retains the complete set of attributes within $x_i$, thereby guaranteeing that $bias_i$ encodes the attributes omitted by $center_i$. The most direct method to maximize this mutual information is to add a reconstruction task. Therefore, when the Valen framework incorporates the Funny method, the overall architecture is as shown in Figure \ref{The Valen framework incorporates the Funny method}.

As shown in this figure, the key attribute representations $\{{center}_i | i \in [1,9]\}$ extracted from the 3$\times$3 image matrix $\{x_i | i \in [1,9]\}$ are superimposed with $\epsilon\sim N(0,1)$ and then fed into the Valen framework for subsequent reasoning. This process uses the Valen model to confirm that representations sampled from $N(center_i,1)$ share the same reasoning identity.
Moreover, Funny introduces three additional regression tasks based on mean squared error (MSE) loss to Valen; these tasks are indicated by dashed lines in red, yellow, and green.
The green dashed line indicates the regression task between the representation of non-key attributes $\{bias_i|i\in[1,9]\}$ and fresh random samples $\epsilon\sim N(0,1)$. 
The objective of this green task is to enforce $bias_i$ to follow $N(0,1)$.
The red and yellow dashed lines indicate two regression tasks performed by the Funny decoder. The red task reconstructs the original RPM image $x_i$ from the combined representation $center_i+bias_i$, while the yellow task reconstructs $x_i$ from $center_i+ \epsilon$.
The objectives of these two reconstruction tasks are vastly different. 
The red task maximizes the conditional distribution $P(x_i|center_i + bias_i)$ during training, which is equivalent to maximizing the mutual information between the attributes of $x_i$ and $center_i + bias_i$. This mutual information can be calculated using the following formula:
\begin{align}
    \text{MI}(x_i, {center}_i + {bias}_i) = H(x_i) - H(x_i | {center}_i + {bias}_i)
\end{align}
where $H(x_i)$ is the entropy of $x_i$, and $H(x_i|center_i + bias_i)$ is the conditional entropy of $x_i$ given $center_i + bias_i$. 
The yellow task forces the output of the Funny decoder to follow the distribution of RPM image pixels, ensuring that the Funny decoder acts as an effective distribution transformation function.
Such MSE-based distribution matching is widely adopted in VAEs \cite{VAE}.

Our design intention for Funny is clear: with the assistance of Valen, Funny compels $center_i$ to store more key attributes related to reasoning, while $bias_i$ accumulates more non-key attributes that could cause interference. Subsequently, the distribution of multi-viewpoint representations produced by $center_i+bias_i$ is planned as a mixture of Gaussians, thereby feeding back to Valen and enhancing its reasoning. The relationship between Funny and Valen is symbiotic. 
Although subsequent experiments demonstrated that Funny is a significantly effective method, given the fact that obtaining disentangled representations under unsupervised learning is not achievable \cite{unsupervised learning of disentangled representations}, this paper holds a pessimistic attitude towards the extent of Funny's decoupling of key and non-key attributes. 

\subsection{Funny not funny}

What is of paramount importance is that, 
a Funny decoder with a conventional structure cannot bear the aforementioned theoretical framework. If we apply a deconvolution structure to the backbone of the Funny decoder, its optimization becomes extremely challenging, and thus hinders satisfactory regression and mutual-information supervision between $center_i+bias_i$ and $x_i$. Alternatively, if we replace the Funny decoder with a ViT structure symmetric to the representation extractor, we obtain excellent regression, yet Valen's normal reasoning accuracy decreases rather than increases.

Unfortunately and ``unfunny", there remains a gap between theory and practice. We speculate that this may arise from the regression task in Funny interfering with the parameter optimization of the reasoning task in Valen.
If we adopt the standard ViT as the Funny decoder, the optimization speed of the regression task would be relatively fast, which could lead to the introduced regression task nearing completion before the reasoning task of Valen. We argue that this premature convergence is the crux: when the regression task approaches completion prematurely, it may lead to $center_i$ falling into the gradient vanishing zone, subsequently rendering it difficult for $center_i$ to make further optimizations for the reasoning task; the non-key attribute stored in $center_i$ can disrupt the reasoning process of Valen. An attempt was made to linearly search for the scaling coefficient between the MSE-based regression task introduced by Funny and the CE-based reasoning task in Valen, but it did not yield an effective solution. 
Anyway, this paper does not intend to bridge this gap between theory and practice by adjusting the scaling coefficients among the various terms in the loss function.

\begin{figure}[htp]\centering
	\includegraphics[trim=0cm 0cm 0cm 0cm, clip, width=8.5
 cm]{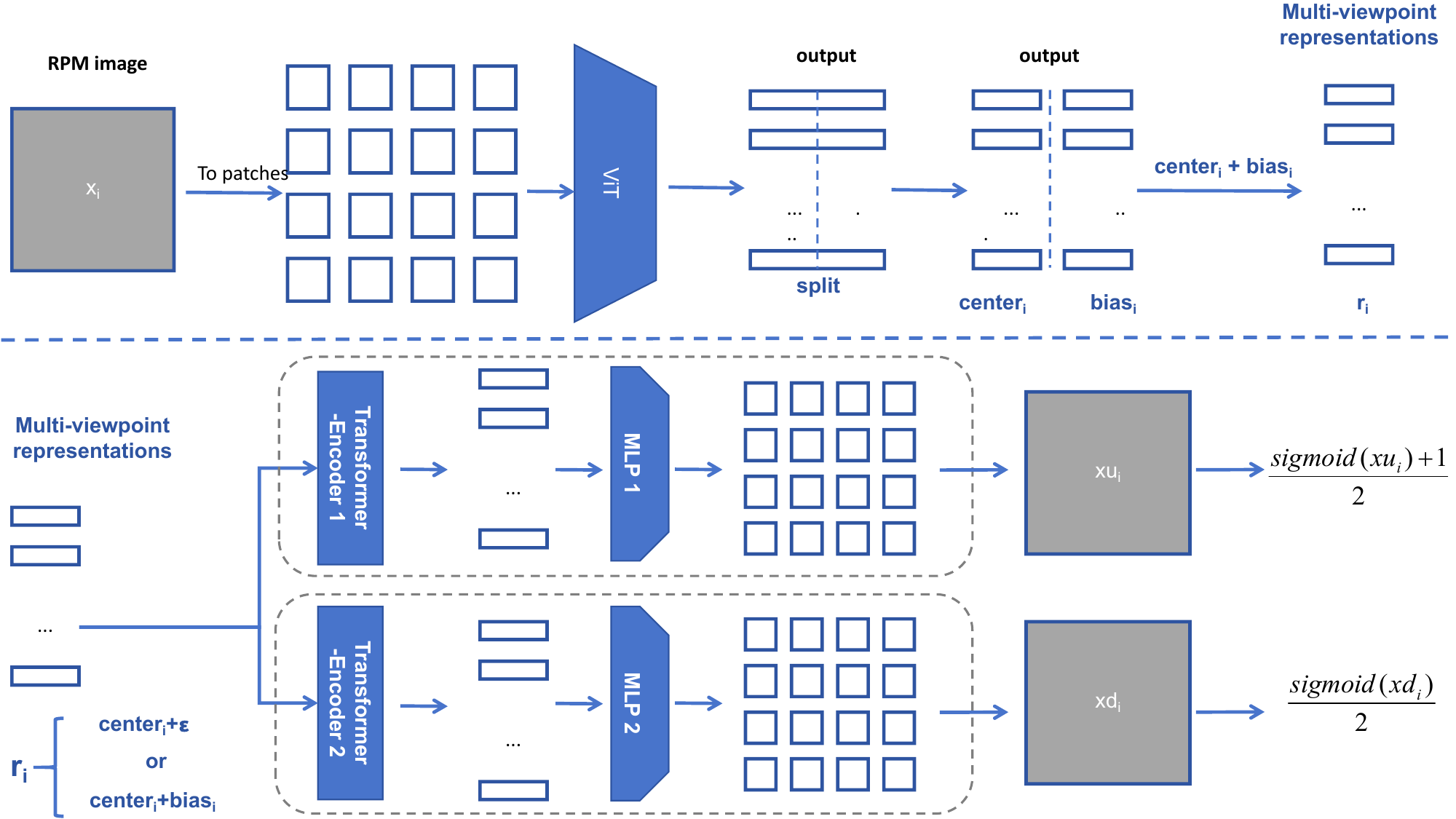}
	\caption{Half-split Decoder}
\label{Half-split Decoder}
\end{figure}

Fortunately and funny, this paper presents a novel decoder architecture, which we term the ``Half-split Decoder" to address this issue. The structure of the Half-split Decoder is illustrated in the Figure \ref{Half-split Decoder}.
The upper half, separated by the blue dashed line, shows Funny’s representation extractor, previously illustrated in Figure~\ref{Funny representation extractor}. The lower half depicts the Half-split Decoder.

As shown in Figure \ref{Half-split Decoder}, the Half-split Decoder is composed of two identical structures, with each structure's parameters optimized independently without parameter sharing. Each branch resembles an inverted version of a ViT\cite{ViT}, which we will not discuss further. The two branches regress the multi-viewpoint representation $r_i$ into two outcomes, $xu_i$ and $xd_i$, respectively. Here $r_i$ can be derived either by summing ${center}_i$ with $ \epsilon$ or by summing ${center}_i$ and $ bias_i$, corresponding to the two regression targets of Funny. 
The core of the Half-split Decoder, however, lies not in these branches but in the two distinct activation functions depicted in Figure \ref{Half-split Decoder}. We express these two activation functions as follows:
\begin{align}
    f_1(xu_i)=\frac{\text{Sigmoid}(xu_i)+1}{2}, f_2(xd_i)=\frac{\text{Sigmoid}(xd_i)}{2}
\end{align}
Clearly, $f_1(xu_i)$  ranges over $(0.5,1)$, while $f_2(xd_i)$ ranges over $(0,0.5)$. Although both ranges differ from the original RPM image pixel range $(0,1)$, we still compute the MSE loss between $f_1(xu_i)$ and $x_i$, as well as between $f_2(xd_i)$ and $x_i$. The sum of these two errors is used as the new loss function for the two regression tasks associated with the Funny decoder. This loss can be expressed as:
\begin{align}
    \ell_\text{Funny}=\text{MSE}(f_1(xu_i),x_i) + \text{MSE}(f_2(xd_i),x_i)
\end{align}
When both $xu_i$ and $xd_i$ are decoded from ${center}_i + {bias}_i$, $\ell_\text{Funny}$ supervises the Mutual Information $\text{MI}(x_i, {center}_i + {bias}_i)$, corresponding to Funny’s red regression task; when both $xu_i$ and $xd_i$ are decoded from ${center}_i + \epsilon$, $\ell_\text{Funny}$ supervises the distribution that the output of the Funny decoder follows, corresponding to the yellow regression task.
The mismatch among the value ranges of \(f_1(xu_i)\), \(f_2(xd_i)\), and \(x_i\) ensures that Funny’s regression loss  $\ell_\text{Funny}$ does not drop to zero during the early stages of Valen's reasoning, thereby preventing the multi-view representation $r_i$ from becoming stagnant.

\subsection{Funny for Bongard-Logo problem}

Funny is essentially a method that modifies the sample representation space based on the key status of sample attributes. Since this key status is context-dependent in Bongard-Logo, the original Funny cannot be directly transferred to this task.
To this end, we argue that identifying a family of unlearnable Bongard-Logo image sample mapping functions $F(x_i)$ that preserve the key attributes under the current context while introducing controlled perturbations to non-key attributes effectively extends Funny’s methodology to this problem. 
Clearly, we retain Funny’s original principle: to explicitly signal to Valen that key attributes determine reasoning identity, while non-key attributes do not affect it.

So, how do we find such functions? Through Human-in-the-Loop (HITL){\cite{HITL}}. HITL refers to situations where human experts or users intervene and provide feedback or guidance during the operation or training of machine learning or deep learning models{\cite{HITL, HITL_sota}}. By analyzing the three coexisting concepts in Bongard-Logo, we found that some straightforward augmentation techniques, such as horizontal or vertical flipping, and 90-degree, 180-degree, and 270-degree rotations, can effectively serve as the function $F(x_i)$. 
Subsequently, to apply Funny to Valen on Bongard-Logo, we simply need to apply these augmentations randomly to the training samples $x_i$ during standard Valen training.
It should be noted that, unlike in Bongard-Logo, the same augmentations can corrupt key attributes in the RPM problem; hence the Funny variant tailored for Bongard-Logo is not transferable to RPM, and vice versa.

\section{SBR:  Method for Planning the Distribution of Reasoning Pattens.}

This paper suggests that we should also plan the distribution of the pattern representation $\{P_{nm}|m\in [1,M]\}_v$ extracted by Valen for visual abstract reasoning problems, so that it aligns with the distribution of the correct solution's pattern.
However, there are no discernible key or non-key attributes in this pattern representation, and it still lacks sufficient interpretability. Therefore, our Funny method is difficult to directly apply to this planning task

To this end, we have designed a new method, SBR (Supervised
Representation Distribution Planning Method). This method can explicitly model the distribution of $\{P_{nm}|m\in [1,M]\}_v$ as a mixture of Gaussian distributions{\cite{GMM}}. 
Its idea is straightforward: we reshape the original distribution of $\{P_{nm}|m\in [1,M]\}_v$ into an interpretable, well-ordered Gaussian mixture and explicitly  supervise the mean and variance of each sub-component in this reshaped distribution.

To establish a mixture of Gaussian distributions, it is necessary to provide means, covariance matrices, and component weights \cite{GMM1}. Consequently, the SBR method needs to specify a reasonable number of components for the distribution of $\{P_{nm}|m\in [1,M]\}_v$, assign appropriate means and variances to these components, and specify the covariance structure. Otherwise, any forced or incorrect specification of the reshaped distribution can degrade Valen's reasoning performance.
It is noteworthy that in RPM problems, each instance includes not only the problem statement, option pool, and index of correct option, but also metadata that provide a detailed description of the progressive patterns inherent in the instance. Interestingly, the intended semantics of $\{P_{nm}|m\in [1,M]\}_v$ encoded by Valen for an RPM instance also corresponds to these progressive patterns.
Therefore, we choose to utilize the metadata to provide detailed information about the target form of the distribution of $\{P_{nm}|m\in [1,M]\}_v$.

\begin{figure}[htp]\centering
	\includegraphics[trim=0cm 0cm 0cm 0cm, clip, width=8.5
 cm]{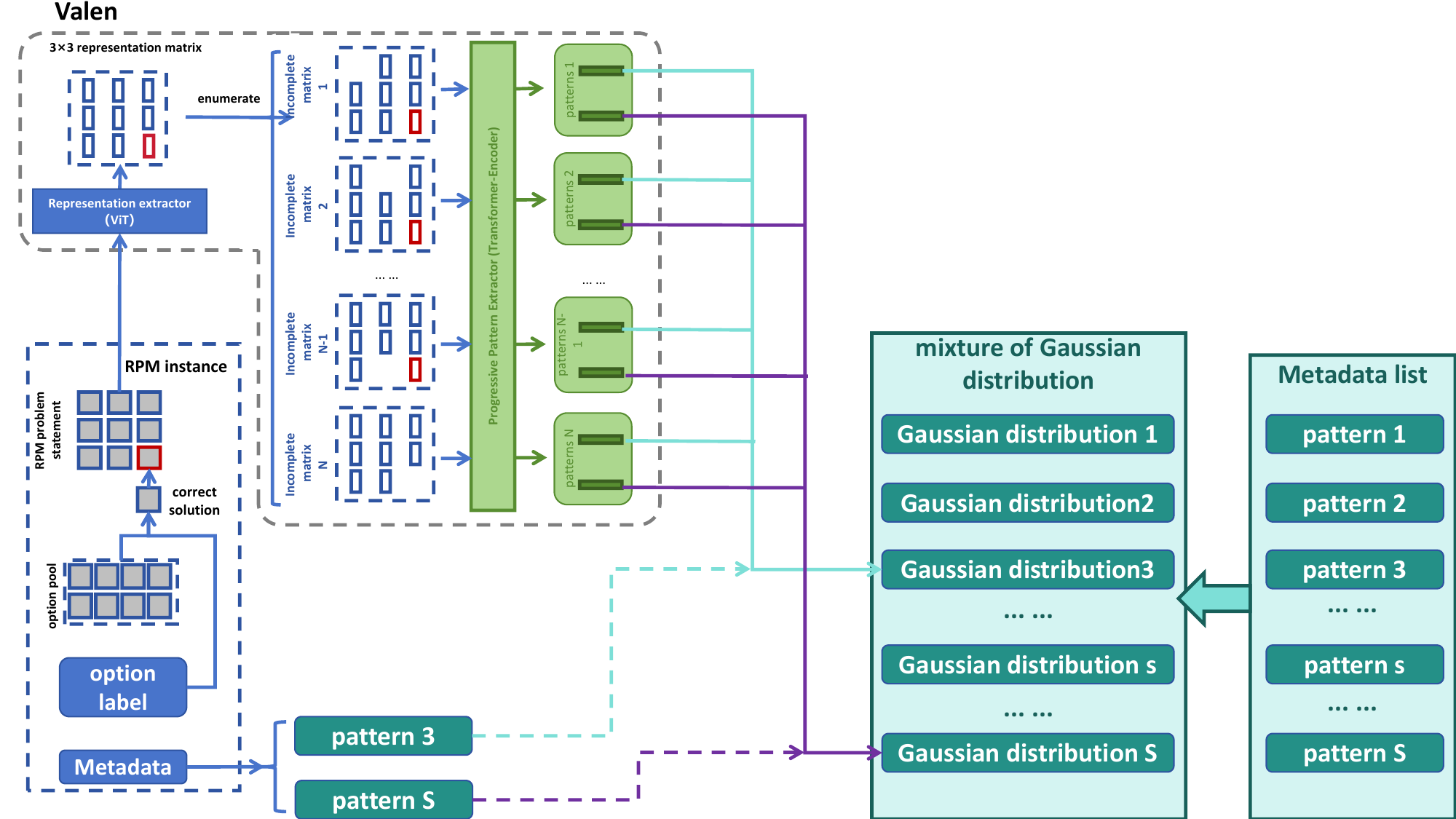}
	\caption{The framework of SBR}
\label{The framework of SBR}
\end{figure}

Specifically, we enumerate all the metadata within RPM instances to obtain a metadata list. After processing each element in the list as a Gaussian distribution, the entire list forms a Gaussian mixture. We then assign the pattern representations $\{P_{nm}|m\in [1,M]\}_v$ to the corresponding component of this mixture, thereby achieving our goal of planning the distribution of pattern representations.
The assigning process can be visualized as Figure \ref{The framework of SBR}.
As shown in this figure, we denote the number of elements in the metadata list to be $S$. 
Since there are often multiple decoupled progressive patterns within the RPM instance, such as the ``shape" and ``line" patterns present in the PGM task, and the accompany metadata fully describe these patterns, we decompose all patterns recorded in the metadata and use them individually to guide the assignment of the $M$ patterns stored in the $\{P_{nm}|m\in [1,M]\}_v$ to their respective distribution components.
Accordingly, we decompose the two progressive patterns, ``pattern 3" and ``pattern $S$", recorded in the metadata and use them separately to guide the distribution attribution of $\{P_{n1}|n\in[1,9]\}_v$ and $\{P_{n2}|n\in[1,9]\}_v$. 
Evidently, when Valen is combined with the SBR method, the value of $M$ is no longer an arbitrary hyperparameter, but must be set to the number of decoupled progressive patterns stored in the metadata; for example, $M = 2$ when the SBR method is applied to the PGM task.

Regarding how we process each element in the enumerated metadata list into distributions, we employ a standard Transformer-Encoder to map the elements, which are descriptions of the progressive patterns, into vectors directly. A list of $S$ elements thus yields $S$ vectors, denoted $\{q_s|s\in [1,S]\}$. By treating the first half of each $q_s$ as the mean and the second half as the logarithm of variance, $\{q_s|s\in [1,S]\}$ defines a Gaussian mixture distribution.
We again use the reparameterization technique shown in Figure \ref{reparameterization} to reparameterize $\{q_s|s\in [1,S]\}$ $T$ times into a new set of vectors, denoted as $\{Q_{st}|s\in [1,S],t\in[1,T]\}$, with $T$ is set to 10.

Regarding the process indicated by the light blue and purple lines in the Figure \ref{The framework of SBR}, where $\{P_{n1}|n\in[1,9]\}_v$ and $\{P_{n2}|n\in[1,9]\}_v$ are separately compressed into the corresponding Gaussian distribution $q_3$ and $q_S$, we achieve this by setting an additional loss function. This loss function can be expressed as follows:
\begin{align}
    \{\bar P_{nm}\}&=\frac{1}{V}\sum_{v=1}^V \{P_{nm}\}_v\label{mean-viewpoint}\\
    {\ell _\text{SBR}}&=  - \sum_{m=1}^M \sum_{\tilde s=1}^S \sum_{n=1}^9 \sum_{t=1}^T \text{meta}_{{m\tilde s}}\cdot \log \frac{{{e^{({\bar P_{nm}} \cdot {Q_{\tilde{s} t}})/\tau}}}}{{ \sum\nolimits_{s = 1}^S {{e^{({\bar P_{nm}} \cdot {Q_{{s} t}})/\tau}}} }}
\end{align}
Where $\text{meta}_{{m\tilde s}}$ indicates whether the the $m$-th progressive pattern recorded in the metadata of current instance matches the $\tilde s$-th element in the metadata list and $\tau$ is a learnable parameter. As shown in Formula (\ref{mean-viewpoint}), to reduce computational overhead we do not compress each viewpoint $v$ of the pattern representations $\{P_{nm}|n\in[1,9], m\in[1,M]\}_v|_{v=1}^V$ individually into the constructed Gaussian mixture; 
instead, we compute the viewpoint average $\{\bar P_{nm}|n\in[1,9], m\in[1,M]\}$ and plan the distribution of these averaged representations.

Formula (\ref{mean-viewpoint}) has clearly elucidated the mechanism of SBR. SBR calculates the probability of each of the $M$ pattern representations in $\{\bar P_{nm}|n\in [1,9],m\in [1,M]\}$ belonging to every component in the mixture of Gaussian distributions $\{q_s|s\in [1,S]\}$, thereby obtaining $M\times S$ probability values. Subsequently, SBR applies the CE loss to constrain the $M\times S$ probabilities according to the indications of metadata. 
These $M\times S$ probabilities are obtained by calculating the sample cosine distance between $\{\bar P_{nm}|n\in [1,9],m\in [1,M]\}$ and $\{Q_{st}|s\in [1,S],t\in[1,T]\}$, and normalizing these distances through the softmax function.

Unfortunately, the descriptions of clustering patterns in Bongard-Logo are overly simplistic, rendering the SBR a specialized solution tailored specifically for RPM problems.


\section{Experiment}

All experiments were programmed in Python using the PyTorch\cite{Pytorch} framework.

\subsection{Experiment on  RAVEN}
We first evaluated Valen on the RAVEN and I-RAVEN datasets. For a fair comparison, we adopted the same settings and equipment as RS-Tran\cite{RS} and Triple-CFN\cite{Triple-CFN}, including data volume, optimizer parameters, data augmentation, batch size, and other hyperparameters. {For the RAVEN problem, the number of viewpoints $V$ in the Valen model is set to 16, corresponding to a $4\times 4$ patch segmentation.}

\begin{table}[h]
\caption{Reasoning Accuracies {Achieved} on RAVEN and I-RAVEN.}
\label{RAVEN_IRAVEN_Results}
\centering
\resizebox{\linewidth}{!}{
\begin{tabular}{cccccccccc}
\toprule
\toprule
&\multicolumn{8}{c}{Test Accuracy(\%)}& \\
\cmidrule{2-9}
Model&Average&Center&2 $\times$ 2 Grid&3 $\times$ 3 Grid&L-R&U-D&O-IC&O-IG \\
\midrule
{CoPINet\cite{CoPINet}}&52.96/22.84&49.45/24.50&61.55/31.10&52.15/25.35&68.10/20.60&65.40/19.85&39.55/19.00&34.55/19.45 \\
\cmidrule{2-9}
{PrAE Learner\cite{PrAE}}&65.03/77.02&76.50/90.45&78.60/85.35&28.55/45.60&90.05/96.25&90.85/97.35&48.05/63.45&42.60/60.70 \\
\cmidrule{2-9}
SAVIR-T \cite{SAVIR-T}&94.0/98.1&97.8/99.5&94.7/98.1&83.8/93.8&97.8/99.6&98.2/99.1&97.6/99.5&88.0/97.2\\
\cmidrule{2-9}
SCL \cite{SCL, SAVIR-T}&91.6/95.0&98.1/99.0&91.0/96.2&82.5/89.5&96.8/97.9&96.5/97.1&96.0/97.6&80.1/87.7\\
\cmidrule{2-9}
MRNet \cite{MRNet}&96.6/-&-/-&-/-&-/-&-/-&-/-&-/-&-/-\\
\cmidrule{2-9}
{DrNet\cite{aaai}}&96.9/97.6&-/-&-/-&-/-&-/-&-/-&-/-&-/-\\
\cmidrule{2-9}
RS-TRAN{\cite{RS}}&{98.4}/98.7&99.8/{100.0}&{99.7}/{99.3}&{95.4}/96.7&99.2/{100.0}&{99.4}/99.7&{99.9}/99.9&{95.4}/95.4 \\
\cmidrule{2-9}
Triple-CFN\cite{Triple-CFN}&98.9/{99.1}&100.0/{100.0}&99.7/{99.8}&96.2/{97.5}&99.8/{99.9}&99.8/{99.9}&99.9/99.9&97.0/{97.3} \\
\cmidrule{2-9}
Triple-CFN+Re-space\cite{Triple-CFN}&99.4/{99.6}&100.0/{100.0}&99.7/{99.8}&98.0/{99.1}&99.9/{100.0}&99.9/{100.0}&99.9/99.9&98.5/{99.0} \\
\cmidrule{2-9}
Valen&99.5/\textbf{99.7}&100.0/\textbf{100.0}&99.7/\textbf{99.8}&98.2/\textbf{99.2}&99.9/\textbf{100.0}&99.9/\textbf{100.0}&99.9/99.9&98.8/\textbf{99.2} \\
\midrule
{Human\cite{CoPINet}}& 84.41& 95.45& 81.82& 79.55& 86.36& 81.81& 86.36& 81.81\\
\midrule
{GPT-4V(0-shot)\cite{LLM2}}& 12.0/-& -/-& -/-& -/-& -/-& -/-& -/-& -/-\\
\cmidrule{2-9}
{GPT-4V(1-shot)\cite{LLM2}}& 12.0/-& -/-& -/-& -/-& -/-& -/-& -/-& -/-\\
\cmidrule{2-9}
{Gemini Pro(0-shot)\cite{LLM2}}& 11.0/-& -/-& -/-& -/-& -/-& -/-& -/-& -/-\\
\cmidrule{2-9}
{Gemini Pro(1-shot)\cite{LLM2}}& 10.0/-& -/-& -/-& -/-& -/-& -/-& -/-& -/-\\
\cmidrule{2-9}
{Genome(few-shot)\cite{LLM3}}& 72.3/-& 80.1/-& -/-& -/-& 67.6/-& 69.1/-& -/-& -/-\\
\bottomrule
\bottomrule
\end{tabular}
}
\end{table}

The experimental records in Table \ref{RAVEN_IRAVEN_Results} demonstrate that existing excellent RPM solvers, including Valen, exhibit varying reasoning accuracy on the RAVEN \cite{RAVENdataset} and I-RAVEN \cite{I-RAVEN} datasets. The I-RAVEN dataset was designed to address limitations in the RAVEN's option pool by enhancing its negative sample setup, such as increasing the number of attributes with shifts and the magnitude of attribute shifts. However, these modifications did not alter the configuration of the primary samples or the distribution of correct solutions. Nonetheless, the same solvers still achieved different reasoning accuracies on RAVEN and I-RAVEN. This observation aligns with our perspective in Tine: the learning objective of probability-highlighting solvers is not the distribution of correct solutions, but rather the distribution delineated by both primary and auxiliary samples.

In Table \ref{RAVEN_IRAVEN_Results}, the performance of several LLMs is listed  below the human benchmark \cite{CoPINet}. Despite high expectations, these models have failed to reach the desired threshold. These findings not only confirm RAVEN’s authority as a benchmark for evaluating AI capability but also demonstrate the superior performance of Valen.

\subsection{Experiment on  Bongard-Logo}

We conducted experiments on Bongard-Logo. {The experiment was conducted in an environment equipped with 4 A100s graphics cards, with a batch size set to 120. The Adam\cite{ADAM} optimizer was chosen, and the initial learning rate was set to $10^{-3}$, with a per-epoch decay method applied for learning rate adjustment, featuring a decay rate of 0.999. It is noteworthy that the Valen model is not sensitive to these parameter settings.}
The results of the experiment are presented in Table \ref{Bongard_Results}. 
\begin{table}[htbp]
\caption{Reasoning Accuracies {Achieved} on Bongard-Logo.}
\label{Bongard_Results}
\centering
\resizebox{\linewidth}{!}{
\begin{tabular}{ccccccc}
\toprule
&\multicolumn{5}{c}{ Accuracy(\%)}& \\
\cmidrule{2-6}
Model&Train& FF&BA&CM&NV \\
\midrule
MetaOptNet\cite{Bongard2} &75.9&	60.3&	71.6&	65.9&	67.5\\
\midrule
ANIL\cite{Bongard2} & 69.7 & 56.6 & 59.0 & 59.6 & 61.0\\
\midrule
Meta-Baseline-SC\cite{Bongard2} & 75.4 & 66.3 & 73.3 & 63.5 & 63.9 \\
\midrule
Meta-Baseline-MoCo\cite{Bongard2} & 81.2 & 65.9 & 72.2 & 63.9& 64.7 \\
\midrule
WReN-Bongard\cite{Bongard2} & 78.7 & 50.1 & 50.9 & 53.8 & 54.3 \\
\midrule
SBSD\cite{PMoC}&83.7&75.2&91.5&71.0&74.1\\
\midrule
PMoC\cite{PMoC}&92.0&90.6&97.7&77.3&76.0\\
\midrule
Triple-CFN\cite{Triple-CFN}&93.2&92.0&{98.2}&{78.0}&{78.1}\\
\midrule
\midrule
Valen&92.5&91.0&97.8&77.5&76.5\\
\midrule
Valen+Tine&{93.3}&{92.4}&{98.2}&{{78.0}}&{77.8}\\
\midrule
Funny+Valen&95.1&93.4&98.8&80.3&79.8\\
\midrule
Funny+Valen+Tine&\textbf{96.0}&\textbf{95.2}&\textbf{99.6}&\textbf{84.6}&\textbf{82.9}\\
\midrule
\midrule
{Human(Expert)\cite{Bongard2}}&-&92.1&99.3&\multicolumn{2}{c}{ 90.7}\\
\midrule
{Human(Amateur)\cite{Bongard2}}&-&88.0&90.0&\multicolumn{2}{c}{71.0}\\
\bottomrule
\end{tabular}
}
\end{table}
As shown in Table \ref{Bongard_Results}, the performance of Funny, Valen and Tine is promising.

Notably, the transformation that adapts the Bongard Logo problem to Valen’s processing framework (illustrated in Figure~\ref{Transformation of Bongard-Logo problem}) is more than merely an RPM-style layout strategy; it is a deliberate redesign grounded in the statistical essence of the Bongard-Logo problem.
From a statistical standpoint, the Bongard-Logo problem involves an unknown distribution \(P({attribute})\) to which the attributes of primary group images conform, whereas those of auxiliary group images do not.
Solving the Bongard-Logo problem hinges on accurately estimating this latent distribution; the better the estimate, the higher the reasoning accuracy.
The Tine method suggests that a probability-highlighting solver is essentially inducing a distribution that assigns high probability to primary group samples and low probability to auxiliary group samples.
When this mechanism is applied to Bongard-Logo, its effect is equivalent to estimating \(P({attribute})\), meaning that if such a solver is deployed on this task, it will inherently focus on that core statistical objective.
Therefore, the fundamental purpose of our transformation of the Bongard-Logo problem is to externalize the core statistical task of estimating \(P({attribute})\) by leveraging the intrinsic mechanism of probability-highlighting solvers, rather than merely providing a convenient interface.

Behind the seemingly simple rearrangement of the reasoning images lies a methodology that externalizes the statistical essence of reasoning problems. The successful transformation of the Bongard-Logo task not only yields a new approach to the task itself but also offers guidance for other reasoning problems whose solutions hinge on estimating attribute distributions.

\subsection{Experiment on  PGM}
In this paper, we conducted experiments related to Valen on PGM. {For the PGM problem, the number of viewpoints $V$ in the Valen model is set to 16, corresponding to a $4\times 4$ patch division. }{For a fair comparison, this paper adopts the same configuration as RS-Tran\cite{RS} and Triple-CFN\cite{Triple-CFN} in terms of hardware selection, data volume, optimizer parameters, data augmentation methods, batch size, and other settings.} And the results are recorded in Table \ref{PGM_Results}. 
\begin{table}[htbp]
\caption{Reasoning Accuracies {Achieved} on PGM.}
\label{PGM_Results}
\centering
\begin{tabular}{ccc}
\toprule
Model&Test Accuracy(\%) \\
\midrule
{CoPINet\cite{CoPINet}}&56.4\\
\midrule
SAVIR-T \cite{SAVIR-T}&91.2\\
\midrule
SCL \cite{SCL, SAVIR-T}&88.9\\
\midrule
MRNet \cite{MRNet}&94.5\\
\midrule
RS-CNN\cite{RS}&82.8\\
\midrule
RS-TRAN\cite{RS}&{97.5}\\
\midrule
Triple-CFN\cite{Triple-CFN}&{97.8}\\
\midrule
Triple-CFN+Re-space layer\cite{Triple-CFN}&{98.2}\\
\midrule
\midrule
Valen&{98.5}\\
\midrule
Valen+Tine&{98.8}\\
\midrule
Funny+Valen&\textbf{{99.3}}\\
\midrule
\midrule
RS-Tran+Tranclip\cite{RS}&99.0\\
\midrule
Meta Triple-CFN\cite{Triple-CFN}&98.4\\
\midrule
Meta Triple-CFN+Re-space layer\cite{Triple-CFN}&99.3\\
\midrule
SBR+Valen&\textbf{99.4}\\
\midrule
\midrule
{Human\cite{PGMdataset}}& $>$80.0\\
\bottomrule
\end{tabular}
\end{table}

From Table \ref{PGM_Results}, it can be observed that the SBR, Tine, and Funny methods have all contributed to varying degrees of improvement in Valen's performance on RPM problems. {The combination of Valen and the Funny method has achieved a reasoning accuracy exceeding 99.3\% on the PGM problem for the first time, without relying on any additional supervisory signals. Furthermore, the combination of SBR with Valen represents the first method that achieves a 99.4\% {reasoning} accuracy by utilizing metadata as an additional supervisory signal. 
The SBR method presented in this paper is not the first to consider utilizing the accompanying metadata within RPM problem instances to assist the solver in reasoning. Previous works\cite{MRNet,RS,Triple-CFN} have explored various approaches to leveraging metadata, but they ultimately found that directly and forcibly imposing additional metadata learning tasks on their solvers was counterproductive\cite{MRNet}. RS-Tran\cite{RS} and Triple-CFN\cite{Triple-CFN} adopted an indirect manner of utilizing metadata. Specifically, RS-Tran utilizes Tranclip\cite{RS} to pre-train its representation encoder using metadata, and Triple-CFN required a similar warm-start process. However, the SBR method in this paper is the first to directly utilize metadata to model the distribution of progressive pattern representations, which is an unexplored approach in previous works and has achieved remarkable results.}

{We also carried out experiments on the generalization issue of PGM, and the findings are documented in Table \ref{Generalization_PGM_2}. As evident from Table \ref{Generalization_PGM_2}, the Funny method significantly enhances Valen's performance on the attribute generalization task Interpolation, whereas SBR exhibits notable advancements in the two progressive generalization patterns, Held-out Pairs of Triples and Held-out Attribute Pairs. }

\begin{table}[h]
\caption{Reasoning Accuracies {Achieved on} PGM Generalization Tasks.}
\label{Generalization_PGM_2}
\centering
\resizebox{\linewidth}{!}{
\begin{tabular}{cccccccc}
\toprule
& \multicolumn{7}{c}{Dataset} \\
\cmidrule{2-8}
\makecell{Model/\\Task} & Interpolation & Extrapolation & \makecell{Held-out\\Attribute\\shape-color} & \makecell{Held-out\\Attribute\\line-type} & \makecell{Held-out\\Triples} & \makecell{Held-out\\Pairs of\\Triples} & \makecell{Held-out\\Attribute\\Pairs} \\
\midrule
\makecell{MRNet\cite{MRNet}} & {68.1} & 19.2 & 16.9 & 30.1 & 25.9 & {55.3} & {38.4} \\
\midrule
\makecell{RS-Tran\cite{RS}} & {77.2} & 19.2 & 12.9 & 24.7 & 22.2 & {43.6} & {28.4} \\
\midrule
\makecell{Triple-CFN+Re-space layer\cite{Triple-CFN} } & {80.4} & 18.4 & 12.6 & 25.2 & 22.0 & {44.5} & {29.2} \\
\midrule
\midrule
Valen & 81.2 & 18.4 & 12.8 & 26.0 & 22.8 & 44.8 & 29.5 \\
\midrule
\makecell{Valen+Tine} & 82.8 & 18.5 & 13.2 & 26.4 & 23.3 & 45.2 & 30.0 \\
\midrule
\makecell{Funny+Valen} & 87.6 & 18.5 & 13.6 & 27.2 & 24.1 & 45.2 & 30.2 \\
\midrule
\makecell{SBR+Valen} & \textbf{92.5} & 12.8 & 13.8 & 26.8 & 28.2 & \textbf{98.1} & \textbf{98.2} \\
\bottomrule
\end{tabular}
}
\end{table}
{
Table \ref{Generalization_PGM_2} also indicates that MRNet demonstrates a slight advantage in the remaining generalization tasks; however, this paper takes a cautious stance, arguing that it is difficult to unequivocally determine which solver is better when the reasoning accuracy hovers around 30\%, as a reasoning accuracy below this threshold already underscores the limitations of the solvers, making it inadequate to discuss superiority at this point.}

\section{Conclusion}
In conclusion, this paper introduces a novel baseline model, Valen, for solving visual abstract reasoning tasks, particularly Bongard-Logo problems and RPM problems. Valen, a probabilistic-highlighting model, demonstrates remarkable performance in handling both image clustering reasoning and image progression pattern reasoning. The paper further delves into the underlying mechanisms of probability-highlighting solvers, realizing that these solvers approximate solutions to reasoning problem instances as distributions. To bridge the discrepancies between the estimated solution distribution and the true correct solution distribution, the paper introduces three methods: Tine, Funny, and SBR. 
\begin{enumerate}
    \item {The Tine algorithm conducts an in-depth analysis of the reasoning mechanism of probability-highlighting solvers and elucidates that the inadequate quality of auxiliary samples in reasoning problem instances often leads to the estimated distribution deviating from the correct solution distribution when using such end-to-end deep learning models to estimate the solution distribution for abstract reasoning problems. Consequently, the Tine method ingeniously incorporates the concept of adversarial learning into the solver's training process, leveraging adversarial techniques to generate more judicious auxiliary samples, for the Valen model, thereby refining its solution distribution estimation process. Tine's contributions lie not only in demonstrating the effectiveness and potential of adversarial learning frameworks in solver training, but also in revealing the crucial role of auxiliary samples in delineating the boundaries of probability-highlighted solver solution distributions, and highlighting the importance of the quality of negative examples for machine intelligence in learning abstract concepts.}
    \item {The Funny method, embracing the spirit of the Tine algorithm, introduces additional reasoning image regression tasks for Valen to achieve mutual information supervision, thereby modeling the distribution of reasoning sample representations encoded by Valen as a Gaussian mixture distribution to enable it to more directly capture the true form of the correct solution distribution. The Half-Split Decoder designed in the Funny method has also made significant contributions to balancing regression and reasoning tasks, potentially opening up new opportunities for solving reasoning problems in the form of Image Generation Model. The Funny method emphasizes the importance of modeling the representation distribution of reasoning samples for enhancing the level of solver's reasoning intelligence.}
    \item {The SBR method further extends the idea of the Funny method. By modeling the distribution of Valen's encoded progressive pattern representations as a more interpretable and orderly mixture of Gaussian distributions, the SBR method significantly enhances Valen's reasoning accuracy and interpretability. The SBR method represents a novel attempt in the field of RPM solver training, embodying an innovative approach and mechanism to utilize metadata. It successfully reverses the previously observed phenomenon where directly introducing metadata as a supervisory signal to the solver actually impairs the reasoning process\cite{MRNet,RS,Triple-CFN}. The SBR method emphasizes that for solvers dedicated to learning abstract concepts, it is crucial to meticulously plan the distribution of latent variables aimed at extracting these abstract concepts, as this significantly enhances their reasoning intelligence.}
\end{enumerate}
 
This paper demonstrates the importance of explicitly planning the distribution of solutions in solving visual abstract reasoning problems.

\newpage

\end{document}


\title{Supplementary Materials for Funny-Valen-Tine: Planning Solution Distribution Enhances Machine Abstract
Reasoning Ability}





\maketitle

\subsection{Ablation study on structure of Valen}  
{
As evident from Figure 7 in the main text, the information flow characteristic of the Valen model lies in the fact that the image information, after being segmented by the representation extractor ViT, is not merged before the final decision. This is because subsequent modules process each viewpoint in parallel and independently. This implies that there is no information exchange during the processing of different viewpoints after the feature extraction module. Consequently, the method of image patches segmentation is particularly crucial. Whether image patches should be segmented based on the entities in the reasoning image to more effectively decouple information, or whether they should be segmented more finely to leverage the advantages of ViT, needs to be verified through experiments.}

{Specifically, in the main text, the Valen model consistently adopts a $4 \times 4$ image patch segmentation method for all RAVEN subproblems. Given Valen's special and effective information flow pattern, we aim to verify the impact on Valen's performance when using corresponding $2 \times 2$ and $3 \times 3$ image patch segmentation methods for the ``$2 \times 2$ Grid'' and ``$3 \times 3$ Grid'' subproblems in RAVEN (or I-RAVEN). The verification results are recorded in Table~\ref{segmentation}.}

\begin{table}[htbp]
\caption{{The Reasoning Accuracy of Valen When Applying a Special Perspective Segmentation Method to the ``2 $\times$ 2 Grid'' and ``3 $\times$ 3 Grid'' Sub-problems in RAVEN and I-RAVEN}}

\label{segmentation}
\centering
\begin{tabular}{cccc}
\toprule
&\multicolumn{2}{c}{ Accuracy(\%)}& \\
\cmidrule{2-3}
segmentation method& 2 $\times$ 2 Grid&3 $\times$ 3 Grid \\
\midrule
Corresponding to the entity&95.2/98.1&93.0/94.5\\
\midrule
Consistently applying the $4 \times 4$&99.7/99.8&98.2/99.2\\
\bottomrule
\end{tabular}
\end{table}

{{The results presented in Table \ref{segmentation} suggest that if Valen segments the reasoning images according to the entity arrangement in the subproblems, its reasoning accuracy will be inferior to that obtained by consistently applying the $4 \times 4$ segmentation method. Furthermore, the results convey an important insight: Valen's special structure allows its multi-viewpoint representation extractor to attain remarkable reasoning accuracy, independent of any prior knowledge about the quantity or position of entities within the reasoning problem.}

\subsection{{Stress test of Valen}}

Unlike PGM and Bongard-Logo problems, the progressive patterns in RAVEN (and I-RAVEN) are guaranteed to exist and are necessarily distributed row-wise. This deterministic nature enables the establishment of a functional correspondence between the number of instances in RAVEN and the number of  progressive patterns embedded within a single instance. This implies that the data volume configuration in RAVEN is not rigidly encapsulated. Consequently, stress tests involving data volume reduction can be conducted on RAVEN.

RS-Tran \cite{RS}, the previous state-of-the-art (SOTA), performed a stress test on three I-RAVEN sub-problems—Center single, 3$\times$3 Grid, and OIG—by halving the training data. We adopt the same split for our stress test; results are shown in Table \ref{The Stress Testing Results}.
\begin{table}[htbp]
\caption{Stress Test on Halved I-RAVEN.}
\label{The Stress Testing Results}
\centering
\begin{tabular}{ccccc}
\toprule
&\multicolumn{3}{c}{ Accuracy(\%)}& \\
\cmidrule{2-4}
Model& Center&3 $\times$ 3 Grid&O-IG \\
\midrule
RS-Tran\cite{RS}&99.6&94.5&95.2\\
\midrule
Valen&99.8&96.8&98.5\\
\bottomrule
\end{tabular}
\end{table}

{Additionally, we conducted a stress test with 80\% of the data removed. The results are recorded in Table \ref{RAVEN_IRAVEN_Results_press}.}
\begin{table}[h]
\caption{Intensive Stress Test on RAVEN and I-RAVEN}
\label{RAVEN_IRAVEN_Results_press}
\centering
\resizebox{\linewidth}{!}{
\begin{tabular}{cccccccccc}
\toprule
\toprule
&\multicolumn{8}{c}{Test Accuracy(\%)}& \\
\cmidrule{2-9}
Model&Average&Center&2 $\times$ 2 Grid&3 $\times$ 3 Grid&L-R&U-D&O-IC&O-IG \\
\midrule
Valen &94.6/95.0&99.0/99.2&92.1/92.4&89.0/89.5&98.0/98.0&97.9/ 97.8&98.1/98.6&90.2/90.3\\
\cmidrule{2-9}
Valen+Tine &95.7/96.1&99.2/99.5&95.0/95.4&89.9/90.5&98.4/98.8&98.3/ 98.6&98.1/98.6&91.0/92.2\\
\cmidrule{2-9}
Funny+Valen&97.2/97.4&99.7/{99.6}&96.7/{97.3}&92.0/{93.4}&99.0/{99.0}&98.9/{99.0}&98.9/99.1&94.4/{94.5} \\
\cmidrule{2-9}
Valen+SBR&98.9/99.0&99.9/{99.9}&99.0/{99.1}&96.6/{97.0}&99.9/{99.9}&99.9/{99.9}&99.9/99.9&97.0/{97.2} \\
\bottomrule
\bottomrule
\end{tabular}
}
\end{table}
{Table~\ref{RAVEN_IRAVEN_Results_press} shows that a sharp reduction in data volume imposes a slight performance penalty on Valen. This decline was expected and is considered acceptable: Transformer architectures are known to be heavily data-dependent in visual tasks, and Valen is built almost entirely upon Transformer blocks. Nevertheless, the Tine and Funny methods largely preserve Valen's performance, and Valen+SBR remains highly competitive, further supporting the utility of these approaches.}

\subsection{Ablation study on structure of Funny decoder}
In this section, we conduct an ablation study on the three regression tasks introduced by the Funny method. The ablation experiment will be conducted on the PGM database. Specifically, we first remove all the regression tasks, and the Valen method combined with Funny degrades to a form as shown in Figure \ref{Ablation_1.pdf}. As can be seen from this figure, in this setting Valen is essentially equivalent to the standard Valen, except that Gaussian noise $\epsilon$ is added to the extracted multi-viewpoint representations. The results of this ablation experiment are recorded in the "Funny 1+Valen" entry of Table \ref{PGM_Results}.
\begin{figure}[htp]\centering
	\includegraphics[trim=3cm 0cm 0cm 0cm, clip, width=8
 cm]{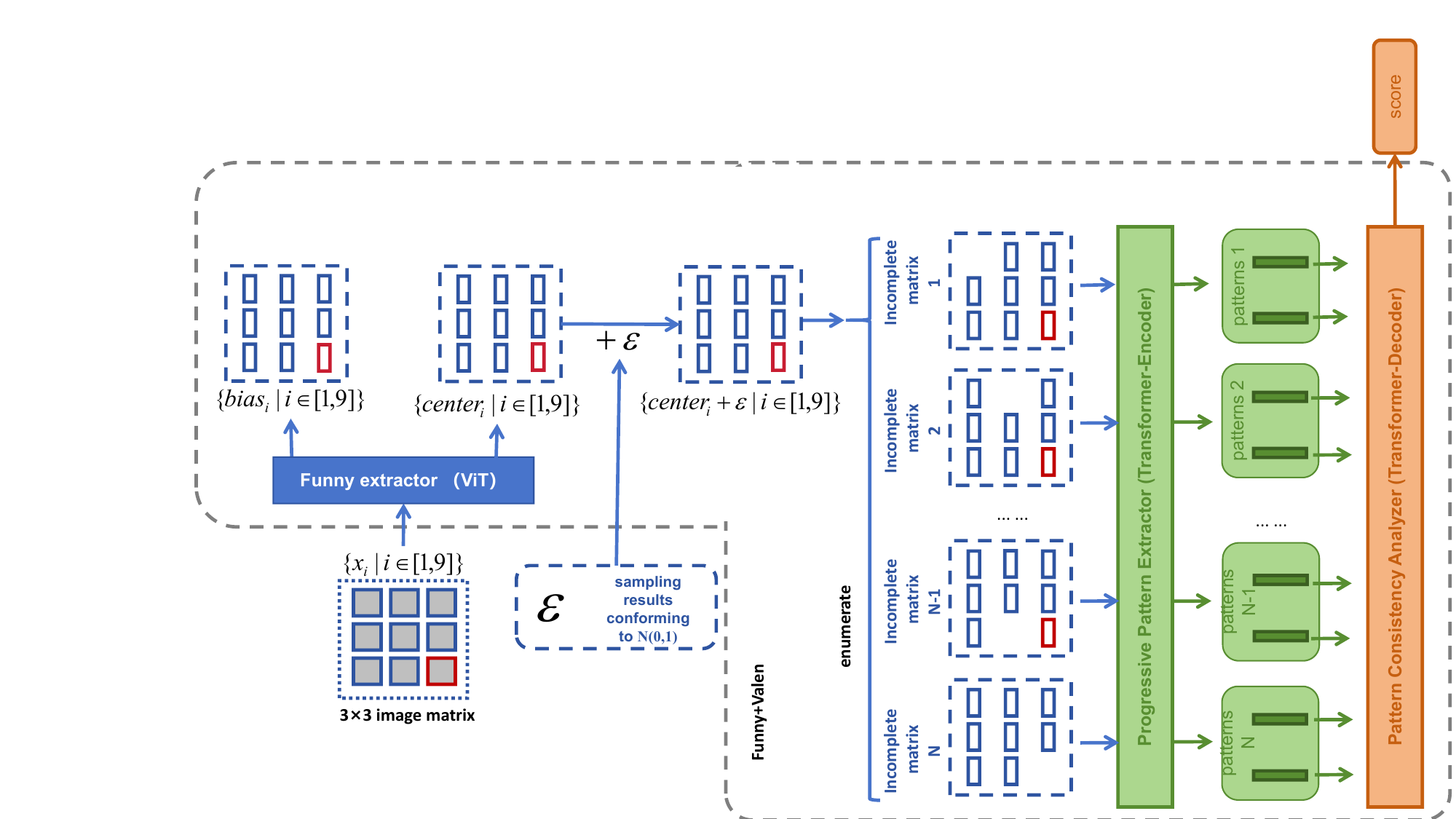}
	\caption{Funny 1+Valen}
\label{Ablation_1.pdf}
\end{figure}

Based on the previous experiment, we reintroduce the regression tasks denoted by the green and yellow dashed lines in Funny, resulting in the combined Funny-Valen model shown in Figure \ref{Ablation_2.pdf}. The results of this Funny configuration on PGM are recorded in the "Funny 2+Valen" entry of Table \ref{PGM_Results}.
\begin{figure}[htp]\centering
	\includegraphics[trim=0cm 0cm 0cm 0cm, clip, width=8
 cm]{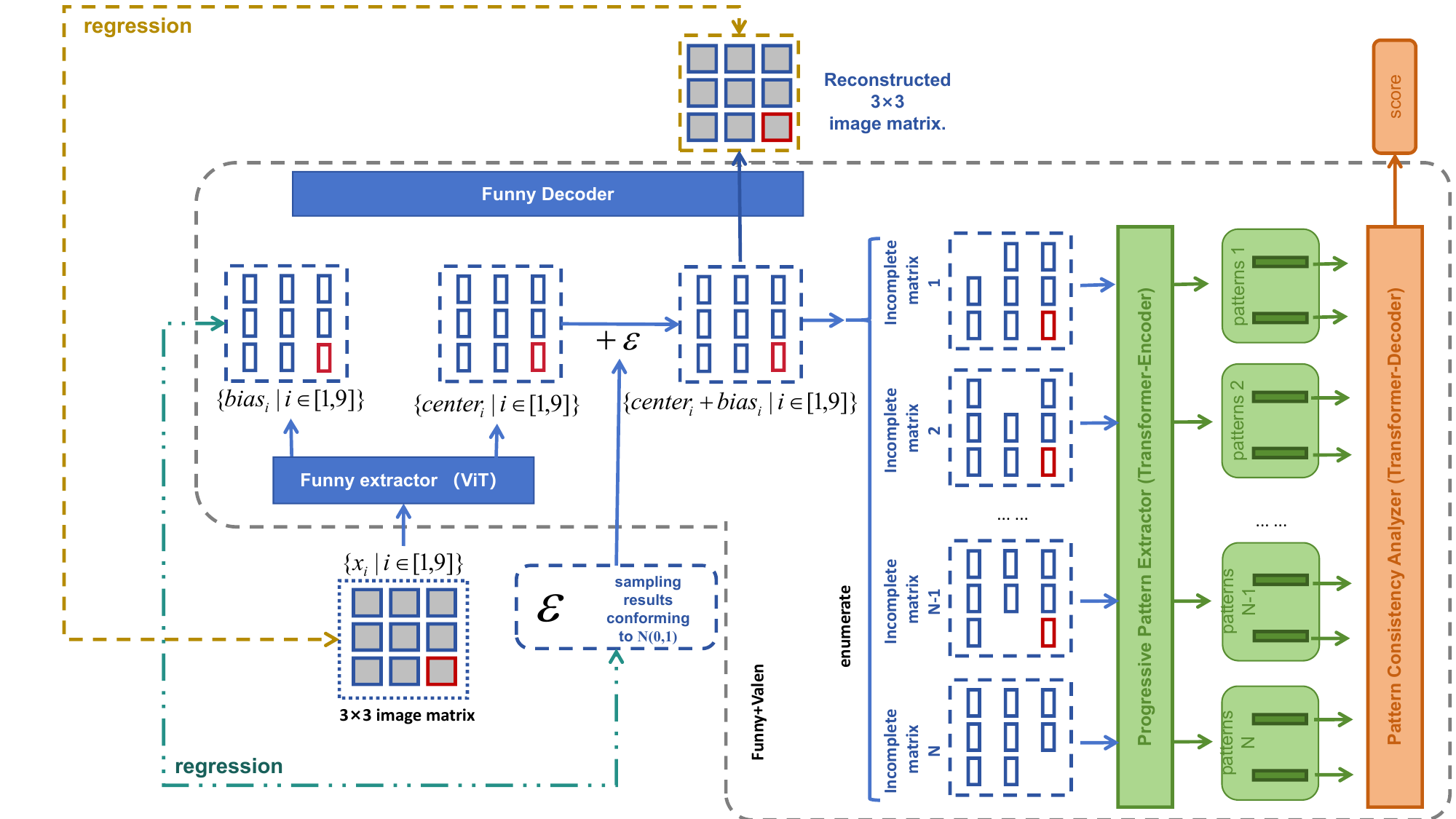}
	\caption{Funny 2+Valen}
\label{Ablation_2.pdf}
\end{figure}

Building on the first ablation, we reinstated the regression tasks denoted by the green and red dashed lines in Funny. The resulting structure is shown in Figure \ref{Ablation_3.pdf}. The results are recorded in the "Funny3+Valen" entry of Table \ref{PGM_Results}.
\begin{figure}[htp]\centering
	\includegraphics[trim=0cm 0cm 0cm 0cm, clip, width=8
 cm]{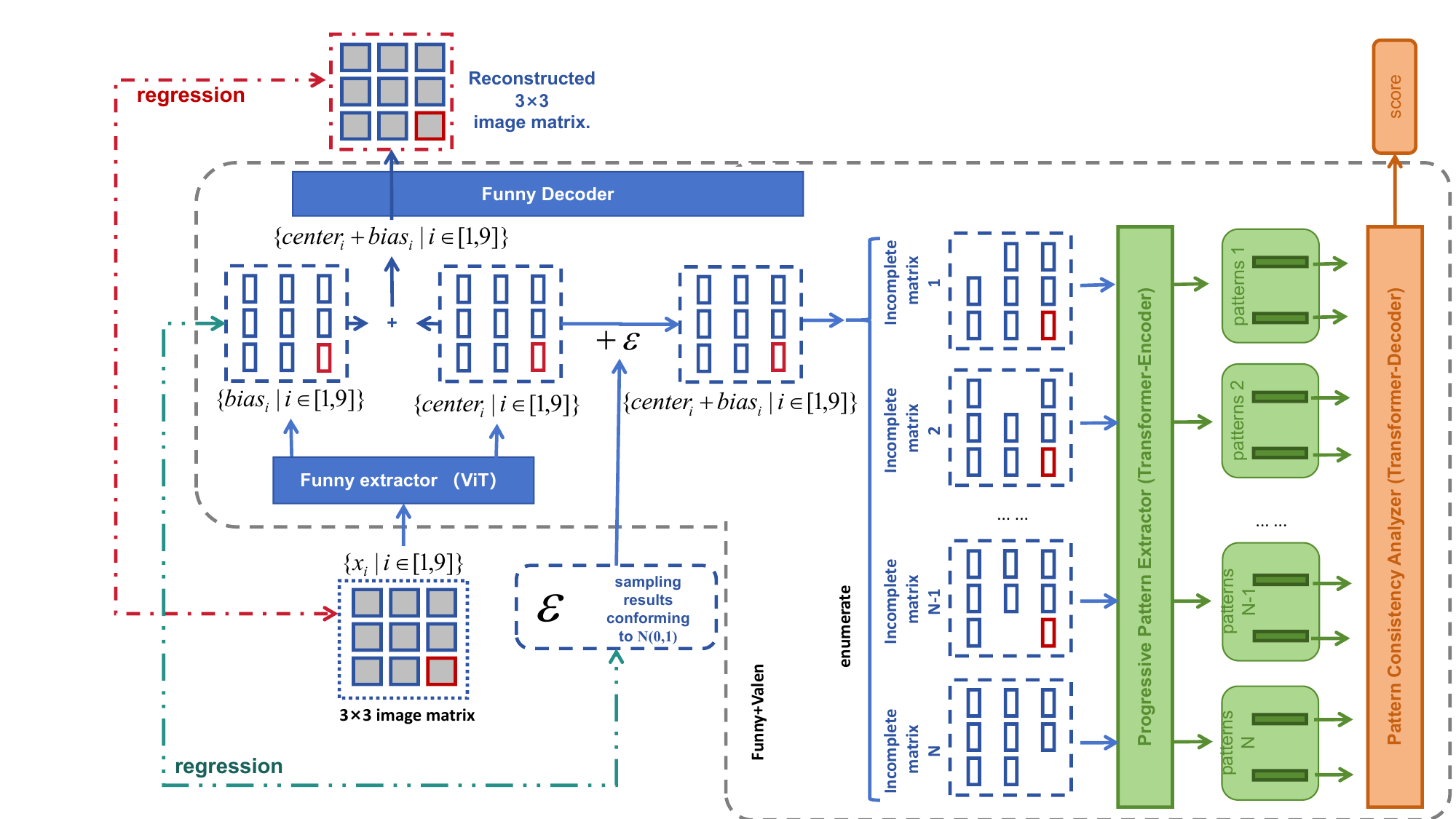}
	\caption{Funny 3+Valen}
\label{Ablation_3.pdf}
\end{figure}

It is worth mentioning that we conducted ablation experiments by replacing the Half-Split Decoder in the Funny method with the standard decoder version of ViT (i.e., inverted ViT), and recorded the corresponding results in Table \ref{PGM_Results} under the entry ``Funny (normal decoder) + Valen". Additionally, we further replaced the Half-Split Decoder in two subsequent ablation studies labeled ``Funny2+Valen" and ``Funny3+Valen". The outcomes of these extended ablation experiments are listed in Table \ref{PGM_Results}, under the headings ``Funny2 (normal decoder) + Valen" and ``Funny3 (normal decoder) + Valen", respectively.

\begin{table}[htbp]
\caption{Reasoning Accuracies of models on PGM.}
\label{PGM_Results}
\centering
\begin{tabular}{ccc}
\toprule
Model&Test Accuracy(\%) \\
\midrule
Valen&{98.5}\\
\midrule
Valen+Tine&{98.8}\\
\midrule
Funny+Valen&\textbf{99.3}\\
\midrule
\midrule
Funny 1+Valen&98.7\\
\midrule
Funny 2+Valen&98.3\\
\midrule
Funny 3+Valen&98.3\\
\midrule
Funny(normal decoder)+Valen&98.4\\
\midrule
Funny 2(normal decoder)+Valen&97.9\\
\midrule
Funny 3(normal decoder)+Valen&97.8\\
\bottomrule
\end{tabular}
\end{table}

\subsection{{The divergence between Valen and humans in interpreting the abstract patterns of Raven's Progressive Matrices}}\label{heatmaps_section}
{This paper argues that it is important to verify whether Valen's interpretation of the abstract patterns embedded in RPM instances aligns with that of humans.
To achieve this purpose, we collected 20 instances each from the ``Center single" and ``$3\times 3$ Grid" sub-problems of the RAVEN dataset. Moreover, we ensured that, from a human perspective, the progressive patterns followed by these 20 instances in each sub-problem were strictly identical. Subsequently, we employed the Valen model, Valen+Tine, Funny+Valen, and SBR+Valen to encode the progressive pattern vectors $\{P_{9m}|m \in[1,M]\}_v$ of these 20 instances one by one. According to the design logic of Valen, $\{P_{9m}|m \in[1,M]\}_v$ represents Valen's interpretation of the progressive patterns of abstract attributes in RAVEN instances. Then, we calculated the cosine similarity between every pair of $\{P_{9m}|m \in[1,M]\}_v$ vectors for these 20 instances and generated a similarity matrix. This enables us to effectively assess whether the progressive patterns that are strictly aligned from a human perspective are also aligned from Valen's perspective.
These similarity matrices were visualized using heatmaps, which are presented in the Figure \ref{heatmaps}.}

\begin{figure}[htp]\centering
	\includegraphics[trim=3cm 1cm 3cm 1cm, clip, width=8.5
 cm]{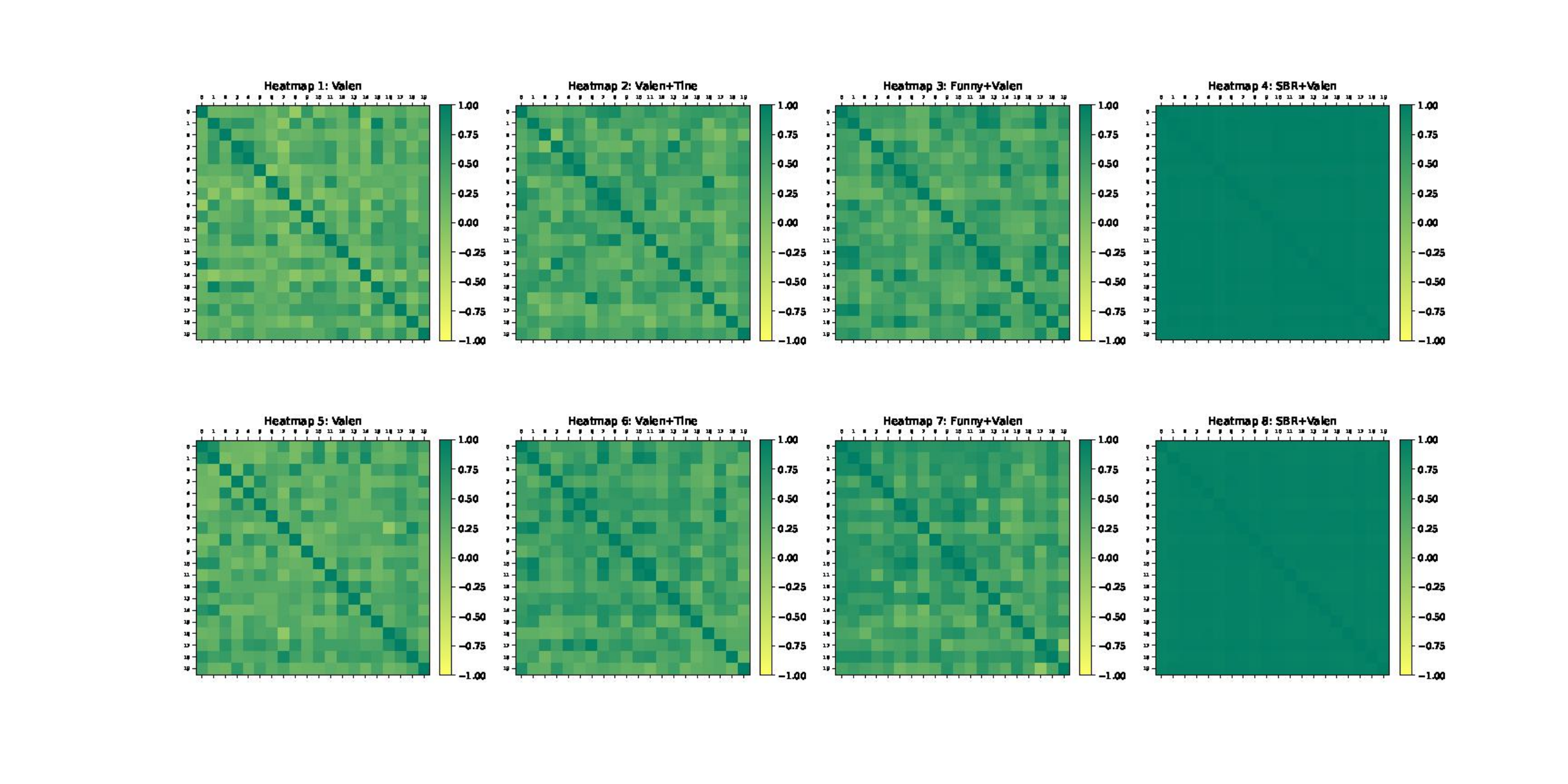}
	\caption{{Heatmaps for validating progressive patterns}}
\label{heatmaps}
\end{figure}


{In figure \ref{heatmaps}, the first four heatmaps are derived from the ``Center single" sub-problem, and the last four are plotted from the ``$3\times 3$ Grid" sub-problem. Heatmap 1 and Heatmap 5 indicate that, despite Valen's structure closely adhering to the problem reasoning logic, its method of inductive reasoning for abstract attributes within RAVEN instances significantly diverges from that of humans. In a measure, this phenomenon corroborates the methodology of Tine. Specifically, the Tine method suggests that the learning objective of a probability-highlighting solver is not the distribution of correct solutions but rather a distribution delineated by both the primary and auxiliary samples. When the auxiliary samples are not judicious, it is foreseeable that the solver is likely to learn a logic that diverges from the problem-solving logic predefined by humans. Thereafter, Heatmap 2, Heatmap 3, Heatmap 6, and Heatmap 7 indicate that the two methods, Tine and Funny, which are employed to refine the solution distribution of the Valen model, have enabled Valen to exhibit a trend of alignment with humans in understanding progressive patterns. It is noteworthy that this divergence precisely served as the design motivation for the SBR method. Heatmap 4 and Heatmap 8 demonstrate that the SBR method, which directly aligns the distribution of $\{P_{nm}\}_v$ with human logic, bridges this divergence, thereby leading to an improvement in both reasoning accuracy and interpretability for Valen.}

\subsection{The interpretability brought by the SBR method for Valen}\label{interpretability_section}

SBR has planned the distribution of $\{P_{nm}|n\in[1,9], m\in [1,M]\}_v|_{v=1}^V$, which enables Valen to exhibit strong human-interpretable characteristics when processing $3\times 3$ progressive matrices. By examining the distribution of $\{P_{nm}\}$, we can inspect Valen's interpretation of the progressive patterns in the current $3\times 3$ matrix. The degree of human-interpretability provided by SBR to Valen can be quantified by assessing the accuracy of Valen's interpretations of matrix progressive patterns. The accuracy of SBR+Valen's interpretations of progressive patterns on PGM is presented in Table \ref{PGM_Pattern_Results}.

\begin{table}[htbp]
\caption{The Accuracy of SBR+Valen's Interpretations of Progressive Patterns on PGM}
\label{PGM_Pattern_Results}
\centering
\begin{tabular}{ccccc}
\toprule
&\multicolumn{3}{c}{ Accuracy(\%)}& \\
\cmidrule{2-4}
Model& shape&line&answer \\
\midrule
SBR+Valen&99.7&99.9&99.4\\
\bottomrule
\end{tabular}
\end{table}

\subsection{{Valen's sensitivity on different attributes.}}\label{attributes_section}

This paper presents an experiment to verify Valen's sensitivity to all key attributes within the RAVEN dataset. RAVEN provides an instance generator whose rule pool can be constrained; using this generator, we produced 1,000 instances with strictly identical progressive patterns for each of RAVEN's two sub-datasets, 3$\times$3 Grid and O-IG, by constraining the rule pool to specific rules. 
Subsequently, we generated another 7,000 instances, but this time replaced the rule pool for a certain key attribute with rules mutually exclusive to the previous ones. As a result, we obtained two sets of generated instances. Compared to the first set, the progressive pattern of that key attribute in the second set was disrupted. These two datasets enable us to validate Valen's sensitivity to the targeted attribute.

The 1:7 ratio between the two datasets naturally gives rise to an eight-option-one-selection validation framework: Valen first extracts the progressive pattern vector group  $\{P_{nm} | n \in [1,9]\}_v$ for all primary samples in both datasets and computes a consistency score for each pattern group within the first dataset; we then draw, without replacement, one $\{P_{9m}\}_v$ from the first dataset and seven $\{P_{nm} | n \in [1,8]\}_v$ from the second dataset, evaluate their pairwise consistencies, and obtain seven scores. Together with the original within-set score, these eight values are compared to reveal Valen’s sensitivity to the targeted attribute.

This paper proposes that the probability of Valen making the correct judgment across these eight consistency scores serves as an effective measure of its sensitivity to the targeted attribute: if Valen is highly sensitive, it will detect the perturbation of that attribute’s progressive pattern and select the correct answer among the eight scores, thereby preserving its reasoning accuracy; conversely, if its sensitivity is low, the same perturbation will be overlooked, leading to an erroneous assessment of the deviation and a consequent drop in accuracy. The results of this comparative analysis are recorded in Table \ref{experimental results of replacing}.


\begin{table}[h]
\caption{{Valen's sensitivity on the Four Key Attributes in I-RAVEN.}}
\label{experimental results of replacing}
\centering
\resizebox{8.5cm}{!}{
\begin{tabular}{lcccccccccc}
\toprule
 & \multicolumn{6}{c}{{Accuracy (Replacing/Not Replacing $P_{nm}$)}} \\
\cmidrule(lr){2-9}
Model & \multicolumn{2}{c}{Attribute:Type} & \multicolumn{2}{c}{Attribute:Size} & \multicolumn{2}{c}{Attribute:Color} &\multicolumn{2}{c}{Attribute:Number/Position}\\
\cmidrule(lr){2-3} \cmidrule(lr){4-5} \cmidrule(lr){6-7} \cmidrule(lr){8-9}
& \makecell{3$\times$3 Grid} & \makecell{O-IG} & \makecell{3$\times$3 Grid} & \makecell{O-IG} & \makecell{3$\times$3 Grid} & \makecell{O-IG} & \makecell{3$\times$3 Grid} & \makecell{O-IG} \\
\midrule
Valen & 99.0/99.1 & 98.9/99.2 & 97.9/99.1 & 98.0/99.2 & 98.5/99.1 & 98.6/99.2 & 98.8/99.1 & 99.0/99.2\\
Valen+Tine & 99.2/99.2 & 99.2/99.3 & 98.1/99.2 & 98.4/99.3 & 98.8/99.2 & 98.8/99.3 & 99.0/99.2 & 99.0/99.3\\
Funny+Valen & 99.2/99.4 & 99.2/99.3  & 98.8/99.4 & 98.8/99.3 & 99.0/99.4 & 98.9/99.3& 99.3/99.4 & 99.3/99.3\\
SBR+Valen & 99.5/99.5 & 99.5/99.5 & 99.5/99.5 & 99.5/99.5  & 99.4/99.5 & 99.4/99.5& 99.5/99.5 & 99.5/99.5\\
\bottomrule
\end{tabular}
}
\end{table}


{Table \ref{experimental results of replacing} shows that Valen and Valen+Tine exhibit varying degrees of sensitivity to the four key attributes. Both demonstrate relatively low sensitivity to the ``Size'' attribute, despite its equal importance to the other key attributes. This further supports the argument in the Tine method that auxiliary samples of inadequate quality may obscure the learning objectives of the solver. However, to some extent, the Funny method can alleviate Valen's relatively low sensitivity to the ``Size'' attribute. Furthermore, the SBR method, which directly manipulates $\{P_{nm}\}_v$, ensures that Valen maintains sufficient sensitivity across all four attributes. These results confirm the effectiveness of both the Funny method and the SBR method.}

\subsection{Cross-task validation}

{Many symbolic models for RAVEN adopt a cross-task validation protocol: they train on one sub-task and then transfer directly to the remaining sub-tasks, using inter-task generalization to assess how robustly the model has captured the underlying progressive patterns. We would like to apply the same diagnostic to Valen. However, unlike Valen’s connectionist pipeline, symbolic approaches treat every RAVEN image as an off-the-shelf bag of entity-attribute values that can be fed straight into reasoning; the cross-task protocol can therefore isolate the robustness of the internalized patterns without perceptual confounds. Connectionist models lack any comparable transparent mechanism, so the protocol cannot disentangle ``pattern-induction robustness'' from ``feature-extraction success'', and hence cannot be applied to Valen unchanged.
To this end, we introduce an equivalent cross-task validation pipeline: first train Valen on a single RAVEN sub-task; then freeze its reasoning module; next realign its multi-viewpoint extractor with only 20\% of the target sub-task’s training data; and finally evaluate on the corresponding test set. The realignment is designed to mimic the entity–attribute priors of symbolic methods, thereby disentangling ``pattern-induction robustness'' from ``feature-extraction success'', so that the cross-task validation focuses on the robustness of the progressive patterns encoded in the frozen module. 
This redesign endows the cross-task protocol with diagnostic validity for Valen.
The cross-task results are in Table \ref{Cross-task validation}. }
\begin{table}[h]
\caption{Cross-Task Accuracy of Valen on RAVEN and I-RAVEN}
\label{Cross-task validation}
\centering
\begin{threeparttable}
\resizebox{\linewidth}{!}{
\begin{tabular}{cccccccc}
\toprule\toprule
& \multicolumn{7}{c}{Target Task Accuracy(\%)}\\
\cmidrule{2-8}
Source Task & Center & 2$\times$2 Grid & 3$\times$3 Grid & L-R & U-D & O-IC & O-IG\\
\midrule
Center &-/-&31.4/31.6&26.4/27.4&39.3/39.4&39.5/39.3&39.5/40.0&39.5/39.9\\
\midrule
3$\times$3 Grid &81.3/83.5&43.9/43.0&-/-&40.2/40.8&40.2/ 40.3&40.0/40.3&40.9/40.4\\
\midrule
O-IG            &79.2/80.0&41.8/42.6&31.8/32.5&41.7/41.3&41.6/41.4&62.0/63.6&-/-\\
\midrule
\midrule
Center$^*$ &-/-&97.8/98.0&60.7/62.0&52.3/52.4&52.5/52.3&52.5/52.0&53.0/53.0\\
\midrule
3$\times$3 Grid$^*$ &99.3/99.5&98.1/99.0&-/-&57.2/57.8&56.2/ 56.3&56.0/56.3&55.9/56.4\\
\midrule
O-IG$^*$  &99.2/99.3&98.0/99.2&83.0/83.5&54.7/55.3&54.6/55.4&99.5/99.6&-/-\\
\bottomrule
\end{tabular}}

\begin{tablenotes}
\footnotesize
\item[ The asterisk ($^*$) denotes the cross-task accuracy obtained by Funny+Valen.]  
\end{tablenotes}
\end{threeparttable}
\end{table}

{As shown in Table \ref{Cross-task validation}, Funny confers a remarkable cross-task performance boost on Valen.
Funny+Valen stably attains approximately 99\% accuracy on two types of transfers: single-track to simplified single-track (3$\times$3$\rightarrow$Center/2$\times$2) and dual-track to compatible single-track (O-IG$\rightarrow$2$\times$2), demonstrating strong downward compatibility. Yet, as soon as the target task increases its track count or modifies attribute constraints, performance collapses:}

{1) Center or 3$\times$3 $\rightarrow$ U-D/L-R/O-IC/O-IG drops to $ 50\%$, presumably because the slot that the reasoning module constructed for single-track patterns cannot accommodate two independent rules at once.}

{2) O-IG $\rightarrow$ U-D/L-R also hovers near $50\%$. We attribute this to mismatched attribute constraints: O-IG's outer entity must fully enclose the inner ones, restricting the size and color of the outer entity, whereas U-D/L-R impose no such restriction, so the internalized slot fails to align. When the restriction of the outer entity is preserved (O-IG $\rightarrow$ O-IC), transfer becomes smooth again, corroborating the hypothesis.}

{3) O-IG $\rightarrow$ 3$\times$3 lands at $ 80\%$, likely because its inner four-entity position-attribute progression is partly incompatible with 3$\times$3's nine-entity counterpart, leaving the reasoning slot imperfectly matched. The large gap between Center $\rightarrow$ 3$\times$3 and Center $\rightarrow$ 2$\times$2 conveys the same message.}

{Compatible patterns are seamlessly adopted, while incompatible ones are actively set aside; both showcase the robustness of Funny+Valen. Overall, its cross-task generalization constitutes a promising step forward for connectionist models.}

\subsection{Image attribute generalization problem within PGM}
The SBR method demonstrates potential in addressing the generalization issue within Probabilistic Graphical Model (PGM) problems. In the context of image attribute generalization problems (Interpolation and Extrapolation) in PGMs, the image attributes present in the training set do not directly appear in the test set. In other words, the progressive patterns of attributes indicated by the metadata in the training set are never encountered in the test set. This results in the supervisory signals used by our SBR method becoming Out-of-Distribution (OOD) supervisory signals at this juncture, which is a discouraging reality. However, we acknowledge that there are numerous methods available for tackling OOD problems, each with its unique advantages, but a common approach is model pre-training. We can leverage SBR to devise a pre-training method to cope with the OOD problem of metadata. Specifically, we first run SBR+Valen on the generalization tasks as usual, then reinitialize the parameters of the Progressive Pattern Extractor and Pattern Consistency Analyzer within the trained Valen, and finally utilize the partially parameter-initialized Valen to relearn the progressive pattern OOD problem independently. The modules that require initialization are outlined with purple dashed lines within the Valen architecture depicted in Figure \ref{Ablation_4.pdf}. The relevant results are recorded in Table \ref{Generalization_PGM_3}.

\begin{figure}[htp]\centering
	\includegraphics[trim=0cm 0cm 0cm 0cm, clip, width=8.5
 cm]{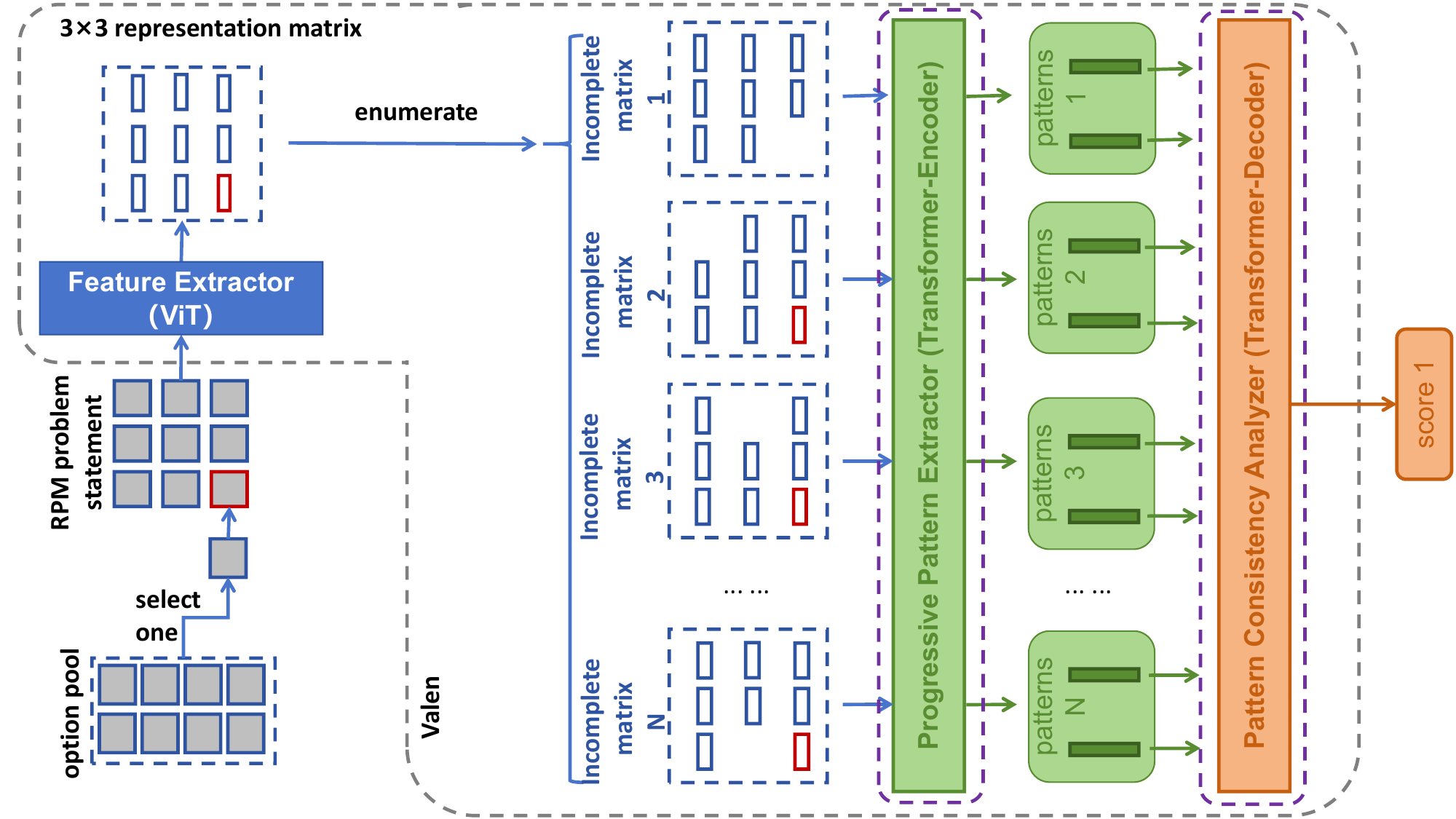}
	\caption{Valen}
\label{Ablation_4.pdf}
\end{figure}



\begin{table}[h]
\caption{Reasoning Accuracies {Achieved on} PGM Generalization Tasks.}
\label{Generalization_PGM_3}
\centering
\renewcommand{\arraystretch}{1.3}
\resizebox{\linewidth}{!}{
\begin{tabular}{cccccccc}
\toprule
& \multicolumn{7}{c}{Accuracy (\%)} \\
\cmidrule{2-8}
\makecell{Model} & Interpolation & Extrapolation & \makecell{Held-out\\Attribute\\shape-colour} & \makecell{Held-out\\Attribute\\line-type} & \makecell{Held-out\\Triples} & \makecell{Held-out\\Pairs of\\Triples} & \makecell{Held-out\\Attribute\\Pairs} \\
\midrule
SBR+Valen & 92.5 & 12.8 & 13.8 & 26.8 & 28.2 & \textbf{98.1} & \textbf{98.2} \\

SBR(pre-train)+Valen & \textbf{94.7} & \textbf{18.6} & 12.8 & 21.4 & 28.1 & 96.2 & 96.9 \\
\bottomrule
\end{tabular}
}
\end{table}
The results indicate that, compared to using SBR to support the entire optimization process of Valen, employing SBR to pre-train the parameters of Valen yields a slight performance improvement in attribute generalization tasks such as interpolation and extrapolation, while experiencing a minor decrease in performance on progressive pattern generalization tasks. These findings hold significant implications for the methodology of utilizing metadata to assist in solver training.

\subsection{{Future work}}
Given that the method proposed in this paper can be effectively integrated into end-to-end solvers, and that large language models (LLMs) also exhibit this end-to-end characteristic, future research on visual abstract reasoning could further explore incorporating solution distribution planning into the training pipeline and architectural design of LLMs, offering new avenues for enhancing their abstract reasoning capabilities.
Meanwhile, we call for future work to include efficiency and parameter count as evaluation metrics, using Valen’s size and speed as baselines to benchmark abstract-reasoning algorithms and foster both better models and sounder standards. Model sizes (MB) are listed in Table \ref{total-parameters-of-models}.
\begin{table}[h]
\caption{Total Parameter Counts for Models}
\label{total-parameters-of-models}
\centering
\resizebox{8cm}{!}{
\begin{tabular}{lcccc}
\toprule
& \multicolumn{4}{c}{{Dataset}} \\
\cmidrule(lr){2-5}
& \makecell{Average \\ (RAVEN \& PGM)} & \makecell{RAVEN} & \makecell{PGM} & \makecell{Bongard-Logo} \\
\midrule
MRNet\cite{MRNet}&6.7&6.7&6.7&-\\
RS-Tran\cite{RS} &3.55&1.0&6.1&-\\
Triple-CFN\cite{Triple-CFN} &7.05&1.7&12.4&54.0\\
PMoC\cite{PMoC}&-&-&-&13.0\\
\midrule
Valen &2.1&2.1&2.1&15.2\\
Valen+Tine &4.7&4.7&4.7&18.9\\
Funny+Valen &16.2&16.2&16.2&18.9\\
SBR+Valen &4.5&4.5&4.5&-\\
\bottomrule
\end{tabular}
}
\end{table}





